\PassOptionsToPackage{table}{xcolor}
\documentclass[sn-basic]{sn-jnl}

\usepackage[table]{xcolor}
\usepackage{graphicx, makecell}
\usepackage{multirow, float,bm,amsmath}
\usepackage{adjustbox}
\usepackage{subcaption}
\usepackage[section]{placeins}

\jyear{2023}%

\theoremstyle{thmstyleone}%
\theoremstyle{thmstyletwo}%

\theoremstyle{thmstylethree}%

\newcommand{\thickhline}{%
	\noalign {\ifnum 0=`}\fi \hrule height 1pt 
	\futurelet \reserved@a \@xhline
}
\newcolumntype{P}[1]{>{\centering\arraybackslash}p{#1}}

\newcommand{\bb}{{\bf b}}

\newcommand{\bx}{{\bf x}}

\newcommand{\bz}{{\bf z}}
\newcommand{\bbw}{{\bf W}}
\newcommand{\bbbw}{{\boldsymbol{\mathcal{W}}}}

\DeclareMathOperator*{\argmax}{arg\,max}

\makeatletter
\def\smallunderbrace#1{\mathop{\vtop{\m@th\ialign{##\crcr
   $\hfil\displaystyle{#1}\hfil$\crcr
   \noalign{\kern3\p@\nointerlineskip}%
   \tiny\upbracefill\crcr\noalign{\kern3\p@}}}}\limits}
\makeatother

\begin{document}

\title[Holistic Deep Learning]{Holistic Deep Learning}


\author*[1]{\fnm{Dimitris} \sur{Bertsimas}}\email{dbertsim@mit.edu}

\author[2]{\fnm{Kimberly} \sur{Villalobos Carballo}}\email{kimvc@mit.edu}

\author[2]{\fnm{Léonard} \sur{Boussioux}}\email{leobix@mit.edu}

\author[2]{\fnm{Michael} \sur{Lingzhi Li}}\email{mlli@mit.edu}

\author[2]{\fnm{Alex} \sur{Paskov}}\email{apaskov@mit.edu}

\author[2]{\fnm{Ivan} \sur{Paskov}}\email{ipaskov@mit.edu}

\affil*[1]{\orgdiv{Sloan School of Management and Operations Research Center}, \orgname{Massachusetts Institute of Technology}, \orgaddress{\city{Cambridge}, \postcode{02139}, \state{MA}, \country{USA}}}

\affil[2]{\orgdiv{Operations Research Center}, \orgname{Massachusetts Institute of Technology}, \orgaddress{\city{Cambridge}, \postcode{02139}, \state{MA}, \country{USA}}}


\abstract{This paper presents a novel holistic deep learning framework that simultaneously addresses the challenges of vulnerability to input perturbations, overparametrization, and performance instability from different train-validation splits. The proposed framework holistically improves accuracy, robustness, sparsity, and stability over standard deep learning models, as demonstrated by extensive experiments on both tabular and image data sets. The results are further validated by ablation experiments and SHAP value analysis, which reveal the interactions and trade-offs between the different evaluation metrics. To support practitioners applying our framework, we provide a prescriptive approach that offers recommendations for selecting an appropriate training loss function based on their specific objectives. All the code to reproduce the results can be found at \url{https://github.com/kimvc7/HDL}.}




\maketitle

\section{Introduction\label{sec:Introduction}}

Neural networks have become increasingly popular due to their remarkable achievements in computer vision and natural language processing. Their generalization power has been demonstrated in wide-ranging applications, from classifying photos to recommending products. However, neural networks face challenges in real-world applications for high-stakes decision-making, including healthcare, policy-making, and autonomous driving. 

First, many standard neural networks are not robust -- they can be easily fooled by natural or artificial noise in the input data \citep{adversarial_robustness}, making them vulnerable to perturbations that may arise in real-world applications. Moreover, neural networks, similar to other machine learning models, often suffer from instability during the training process -- different train-validation splits could generate models with very different performance \citep{may2010data, xu2018splitting}. This reduces the policymakers' trust in these models and hinders post-hoc interpretations. Another critical difficulty is that neural networks are not sparse -- the high number of parameters utilized for neural networks prevents efficient computation and storage \citep{DBLP:journals/corr/abs-2007-05558}. Most neural networks have millions of non-zero parameters to be stored and accessed for evaluation. This is problematic in many decision-making settings with limitations or restrictions on hardware capabilities. Reducing the number of parameters could make them more applicable in a broader range of scenarios \citep{changpinyo2017power, narang2017exploring}.

The questions around improving robustness, stability, and sparsity metrics have all been previously studied in the neural network literature. However, they have been almost exclusively studied in isolation, with a limited understanding of the tradeoffs between these desired qualities. This paper aims to simultaneously address all these objectives through a novel comprehensive methodology named Holistic Deep Learning (HDL). In particular, HDL carefully combines state-of-the-art techniques that address these individual challenges and demonstrates their collective efficacy through extensive experiments on diverse data sets. Our findings provide a promising pathway toward developing efficient and reliable machine learning models across many dimensions for real-world applications.\\

\noindent Specifically, our contributions are as follows: 
\begin{enumerate}
\item We design HDL, a novel framework that jointly optimizes for neural network robustness (adversarial accuracy), stability (worst accuracy across train-val splits), and sparsity (parameters with value zero) metrics by appropriately modifying the objective function.
\item Through extensive ablation experiments and SHAP value analysis \citep{lundberg2017unified} across 45 UCI data sets \citep{uci} and 3 image data sets (MNIST \citep{mnist}, Fashion MNIST \citep{fashionmnist} and CIFAR10 \citep{cifar}), we analyze the individual performance of each metric as well as the interactions and trade-offs between them. We corroborate that imposing robustness, stability, and sparsity improves the corresponding metrics across all data sets. In addition, we show that:
\begin{itemize}
    \item imposing stability and sparsity further improves robustness,
    \item imposing stability and robustness further improves sparsity,
    \item imposing robustness further improves stability,
    \item imposing stability and robustness further improves natural accuracy.
\end{itemize} 
The effect of sparsity on natural accuracy is more complex and highly varies across data sets. However, we show that it is often possible to simultaneously improve robustness, stability, and sparsity without sacrificing performance on natural accuracy.
\item We propose a prescriptive approach to provide recommendations on selecting the appropriate loss function depending on the practitioner's objective. In particular, simultaneously imposing robustness, stability and sparsity in the loss function leads to the best results when jointly optimizing for all the metrics. 

\end{enumerate}

The paper is organized as follows: Section \ref{sec:related} outlines the current literature of robust, sparse, and stable methods; 
Section \ref{sec:methodology} describes the Holistic Deep Learning framework, and Section \ref{sec_comp_exp_regr} shows the results of the computational experiments.

\section{Related Work} \label{sec:related}


\subsection{Robust Neural Networks}

Many state-of-the-art deep neural networks are highly vulnerable to small perturbations in the input data~\citep{adversarial_robustness}, which can be a threat to some real-world applications like self-driving cars or face recognition. Adversarial robustness evaluates a neural network's resistance against these altered inputs, referred to as adversarial examples, intentionally designed to worsen the network's performance \citep{adverse_goodfellow, adverse_carlini, madry}. 

Multiple methods have been developed in recent years to enhance the adversarial robustness of neural networks. One of the most popular heuristics is augmenting the data set during training with adversarial examples~\citep{madry, adverse_goodfellow}. Others include neuron randomization~\citep{prakash, xie2017mitigating}, input space projections~\citep{lamb, kabilan, samangouei} and regularization \citep{DBRobustness, ross2017improving, DBLP:journals/corr/HeinA17, yan2018deep}. A less common but more theoretically rigorous approach is to minimize a provable upper bound of the loss achieved with adversarial examples~\citep{raghunathan2018semidefinite,singh2018fast,zhang2018efficient,weng2018towards,dvijotham2018dual,lecuyer2019certified,cohen2019certified,anderson2020strong,DBRobustness}.

While these methods successfully improve the network's robustness, the extent to which they do so often depends on the data set, the network size, and the magnitude of the input perturbations. In particular, heuristic methods generally work well for small perturbations, while the upper bound methods yield better results when the input noise is larger~\citep{DBRobustness, athalye2018obfuscated}. However, there is a trade-off between effectiveness and efficiency. The methods providing the strongest adversarial robustness are often computationally demanding, making it challenging to implement them for large data sets or complex network architectures.

\subsection{Sparse Neural Networks}

In machine learning, sparse models make predictions based on a limited number of parameters. Sparsity is often desirable, as it may save memory, enhance model interpretability, and reduce overfitting \citep{sparse-reg}. 

There are two typical approaches to sparsity in deep learning. The first one, \textit{train-then-sparsify}, consists of removing unnecessary neurons or connections after training the network, sometimes followed by retraining~\citep{janowski,lecun}. This approach has been widely investigated, and several schemes exist to choose which connections to prune~\citep{sparsityindeep}. \cite{han_sparse}, for example, propose to prune the connections with the smallest weights. Other methods include formulating a convex optimization problem~\citep{aghasi}, removing filters for which the total absolute sum is low~\citep{li}, and eliminating channels that have limited impact on the network's discriminatory ability~\citep{zhuang}. The second approach, \textit{sparsify-during-training}, is achieved by learning a sparse architecture while training the network. Multiple methodologies exist \citep{bellec, mocanu, mostafa}, including the method to approximate the $\ell_0$ norm with continuous functions and add a regularization term to the loss function \citep{louizos2018learning, savarese2020winning}. We refer the reader to \cite{stateofsparsity} and \cite{sparsityindeep} for more comprehensive surveys on sparsity. 

\subsection{Stable Neural Networks}
\label{sec:stabledef}

The stochastic nature of data samples can lead to instability and high dependence of machine learning models on the specific train-validation split. This can negatively impact the interpretability of the resulting model and its ability to make reliable predictions \citep{StableRegression}, a key factor to establishing trust in any algorithm.

The sensitivity of machine learning models to the choice of training split has mostly been studied through the lens of cross-validation and distributionally robust optimization. Cross-validation can be used to measure the variability from the selection of training split but at a significant increase in computational cost \citep{CV1,ESL} that is often intractable for deep learning settings. Distributionally robust optimization has been used to quantify the worst-case generalization error in the presence of shifts in distribution or regime \citep{DR1, DR2, pangwei}, but it often requires pre-defined groups over the training data and expensive group annotations for each data sample to avoid overly pessimistic uncertain distributions \citep{pangwei, liu2021just}. A different approach has been studied by \cite{StableRegression} and \cite{StableClassification}, who instead optimize over the worst training set of fixed size without making any probabilistic assumptions. Although their method was presented in the context of linear and tree-based models, their framework also applies to neural networks.

\section{The Holistic Deep Learning Approach}
\label{sec:methodology}
\subsection{The HDL Framework}
\label{subsec:hdl-network}
We introduce the HDL framework for a classification problem over points $\bx\in \mathbb{R}^M$ whose target $y\in [K]$ is one of $K$ different classes (we use the notation [n] to denote the set $\{1, \hdots, n\}$). We illustrate our approach over a fully-connected neural network for simplicity of notation, but the framework remains the same for convolutional neural networks.

For $x \in \mathbb{R}$, define $[x]^+ = \max \{0,x\}$ (the ReLU function). Given weight matrices $\bbw^\ell\in \mathbb{R}^{r_{\ell-1}\times r_{\ell}}$ and bias vectors $\bb^\ell\in \mathbb{R}^{r_\ell}$ for $\ell\in [L]$,  such that $r_0= M, r_L = K$, the corresponding feed-forward neural network with $L$ layers and ReLU activation functions is defined by the equations:
\begin{align}
   \bz^1(\theta, \bx) &= \bbw^1\bx+\bb^1,\\
    \bz^\ell(\theta, \bx) &= \bbw^\ell [\bz^{\ell-1}(\theta, \bx)]^+ + \bb^{\ell}, \quad \forall \: 2\leq \ell \leq L, \label{eq:key12}
\end{align}
where $\bbbw$ denotes the parameters $(\bbw^\ell, \bb^\ell)$ for all $\ell \in [L]$. Consider a data set $\{({\bx}_n, {y}_n)\}_{n=1}^N$, where $y_n\in[K]$ is the target class of $\bx_n$. For each point $\bx_n$, the class predicted by the network is $ \argmax_k \: {z}^L_k(\bbbw, \bx_n)$. 

The nominal DL approach is to minimize the cross-entropy loss of the network $\bz^L$ described in Eq. \eqref{eq:key12}, which can be written as:
\begin{equation}
\min_{\bbbw} \frac{1}{N}\sum_{n=1}^N \log\left(\sum_{k = 1}^K e^{ (\Delta \boldsymbol{e}_k^{y_n})^\top{\bz}^L(\mathcal{\bbbw}, \bx ) }\right),
\end{equation}
where $\Delta \boldsymbol{e}_k^{y_n} = \boldsymbol{e}_k - \boldsymbol{e}_{y_n}$ and $\boldsymbol{e}_k$ refers to the one-hot vectors with a $1$ in the $k^{\text{th}}$ coordinate and $0$ everywhere else. In our HDL framework, we propose instead to minimize the following optimization problem: 
\begin{equation}
\begin{aligned}
\min_{\bm{s}, \theta, \bbbw} &\lambda \underbrace{\sum_{j=1}^{\vert\bbbw\vert}  
\sigma \left(\beta s_j \right)}_{\text{Sparsity}}+ \underbrace{\theta}_{\text{Stability}} + \\
\frac{1}{a}&\sum_{n=1}^N \left[\log{\sum_{k=1}^K e^{(\Delta \boldsymbol{e}_k^{y_n})^\top{\bz}^L(\bbbw \odot \sigma(\beta \bm{s}), \bx ) + \overbrace{\footnotesize{\rho \|\nabla_{\bx} (\Delta \boldsymbol{e}_k^{y_n})^\top{\bz}^L(\bbbw \odot \sigma (\beta \bm{s}), \bx)\|_1}}^{ \text{Robustness}}}} -\hspace{-0.8cm} \smallunderbrace{\theta}_{\text{Stability\hspace{0.7cm}}}\hspace{-0.7cm}\right]^+\hspace{-0.3cm},\label{eq:hdl}
\end{aligned} 
\end{equation}
where $\odot$ corresponds to the element-wise product, $\sigma$  is the standard sigmoid function, ${\bz}^L(\cdot, \bx)$ was defined in \eqref{eq:key12}, $\lambda$ (resp. $\rho)$ is the regularization coefficient corresponding to the sparsity (resp. robustness) loss component, and $a$ is the size of the data subsets used for the stability requirement (see Section \ref{sec:stabledef}). We observe that robustness adds a term to the output, while stability and sparsity add new parameters ($\theta$ and $\bm{s}$ respectively) to be optimized. This loss function allows us to simultaneously train robust, sparse, and stable feed-forward neural networks at scale. In the next section, we provide more details about each metric.

\subsection{Robustness}

This section describes our method to introduce the robust component into neural network training. Since our ultimate goal is to incorporate the sparsity, robustness, and stability of neural networks together in a tractable way, we avoid training algorithms that significantly increase the training time or the algorithm's complexity. We follow the approach from \cite{DBRobustness} of using a linear approximation of the neural network to estimate the robust objective. This approach is simple to implement, produces good adversarial accuracy, and does not require the extensive training time of other algorithms.

For a given $(\bx, y)$ pair, the robust problem using the cross-entropy loss and the $\ell_{\infty}$-norm uncertainty sets can be upper bounded as: 
\begin{equation}
\begin{aligned}
\max_{\boldsymbol{\delta}: \|\boldsymbol{\delta}\|_{\infty}\leq \rho}\log\sum_{k} e^{(\Delta \boldsymbol{e}_k^{y})^\top{\bz}^L(\bbbw, \bx + \boldsymbol{\delta})}  \leq \log \sum_{k} e^{\max_{\boldsymbol{\delta}: \|\boldsymbol{\delta}\|_{\infty}\leq \rho} (\Delta \boldsymbol{e}_k^{y})^\top{\bz}^L(\bbbw, \bx + \boldsymbol{\delta})}.
\label{eq:upper-bound}
\end{aligned}
\end{equation}
Since ${z}^L_k(\mathcal{W}, \bx)$ is piece-wise linear, we expect the outputs ${z}^L_k(\bbbw, \bx)$ and ${z}^L_k(\bbbw, \bx + \boldsymbol{\delta})$ to be in the same linear piece when $\bx + \boldsymbol{\delta}$ is close to $\bx$. In other words, the linear approximation 
\begin{align}
    {z}^L_k(\bbbw, \bx + \boldsymbol{\delta}) \approx {z}^L_k(\bbbw, \bx) +  \boldsymbol{\delta}^\top \nabla_{\bx} {z}^L_k(\bbbw, \bx) 
\end{align}
 is exact for small enough $\boldsymbol{\delta}$. Therefore, we approximate the upper bound in \eqref{eq:upper-bound} as
\begin{align}
    &\log\sum_{k} e^{\max_{\boldsymbol{\delta}: \|\boldsymbol{\delta}\|_{\infty}\leq \rho} (\Delta \boldsymbol{e}_k^{y})^\top{\bz}^L(\mathcal{W}, \bx + \boldsymbol{\delta})}\nonumber \\
     \approx &\log\sum_{k} e^{\max_{\boldsymbol{\delta}: \|\boldsymbol{\delta}\|_{\infty}\leq \rho} (\Delta \boldsymbol{e}_k^{y})^\top{\bz}^L(\mathcal{W}, \bx ) + \boldsymbol{\delta}^\top \nabla_{\bx} (\Delta \boldsymbol{e}_k^{y})^\top {\bz}^L(\bbbw, \bx)} \nonumber \\
     = &\log\sum_{k} e^{ (\Delta \boldsymbol{e}_k^{y})^\top{\bz}^L(\mathcal{W}, \bx ) + \rho \| (\Delta \boldsymbol{e}_k^{y})^\top\nabla_{\bx}{\bz}^L(\bbbw, \bx)\|_1}.
     \label{eq:approx-upper-bound}
\end{align}
Even though the expression in Eq. \eqref{eq:approx-upper-bound} is not always an upper bound of Eq. \eqref{eq:upper-bound} for an arbitrary value of $\rho$,  \cite{DBRobustness}  experimentally show that generally the average loss obtained using this expression is indeed an upper bound of the average adversarial loss. In fact, for small $\rho$, their experiments demonstrate that this approach achieves competitive results with state-of-the-art methods while requiring significantly less computational time across various tabular and image data sets. However, we emphasize that the methodology developed in this paper could also be performed with other methods for robust training, like adversarial training or upper bound minimization, which might be more appropriate for large uncertainty sets. 

\subsection{Sparsity} \label{sec:sparsity}


In this work, we use the specific retraining procedure proposed by \cite{savarese2020winning}, which deterministically approximates the $\ell_0$ regularization utilizing a sequence of sigmoid functions and adding them as a penalty term in the loss function. Notably, the implementation is easily compatible with our robustness and stability requirements, since this methodology relies on a penalty term added in the loss function. 
Therefore, we can use gradient descent to simultaneously optimize the objective function comprising the robustness, stability, and sparsity penalties. 

Adding $\ell_0$ regularization explicitly penalizes the number of non-zero weights in the model to induce sparsity. However, the $\ell_0$-norm induces a priori a non-convex and non-differentiable loss function $\mathcal{R}(\boldsymbol{\bbbw})$, as follows:
\begin{equation}
\mathcal{R}(\boldsymbol{\bbbw})=\frac{1}{N}\left(\sum_{n=1}^{N} \mathcal{L}\left(y_{n}, {\bz}^L(\bbbw, \bx_n)\right)\right)+\lambda\|\bbbw\|_{0}, \quad\|\bbbw\|_{0}=\sum_{j=1}^{\vert\bbbw\vert} \mathbb{I}\left[w_{j} \neq 0\right],
\end{equation}
where $\vert\bbbw\vert$ is the number of parameters, $w_j$ is the $j^{th}$ coordinate of $\bbbw$, $\lambda$ is the regularization weight and $\mathcal{L}$ a loss function (e.g., cross-entropy loss). 

The goal is to relax the discrete nature of the $\ell_0$ penalty to preserve an efficient continuous optimization while allowing for exact zeros in the neural network weights. To do this, \cite{savarese2020winning}  propose to first parameterize the weights $w_j=H(s_j)$ where $H(\cdot)$ is the Heaviside step function, and then approximate the non-differentiable step function with the sigmoid function: $\sigma(\beta s_j) \to H(s_j)$ when $\beta \to \infty$. Therefore, $\beta$ is the hardness parameter that controls how close the approximation is to the $\ell_0$ regularization, and the final loss function can be written as:
\begin{equation}\label{eq:l0}
\mathcal{R}(\boldsymbol{\bbbw})\approx \frac{1}{N}\left(\sum_{n=1}^{N} \mathcal{L}\left(y_{n}, {\bz}^L(\bbbw \odot \sigma (\beta \bm{s}), \bx_n)\right)\right)+\lambda\sum_{j=1}^{\vert\bbbw\vert} \sigma(\beta s_j).
\end{equation}
To achieve a sparse network, we use this loss function (\ref{eq:l0}) over multiple training rounds to gradually reach a sparse initialization before training the final sparse neural network. To obtain each initialization before a new training round, we start with our initialized auxiliary sparsity $\bm{s}_0$ and hardness $\beta=1$ parameters. Over the $T$ training iterations, we gradually increase $\beta$ until it reaches a maximum value $\bar{\beta}$ when the training procedure is completed with sparsity $\bm{s}_{T}$. Then, we take $\bm{s}_0'=\min(\bar{\beta} \bm{s}_T, \bm{s}_0)$ to generate the new initialization for the next round of training. This minimization function essentially keeps the information of the suppressed weights, i.e., $\sigma(\beta s_j) \approx 0$, while reverting those not suppressed to their starting position. This process is completed over multiple rounds to find better and sparser initializations for the neural network.

We implement the methodology as suggested by \cite{savarese2020winning}. In the results section, we measure sparsity in terms of the percentage of neuron connections (weights) set to 0.

 \subsection{Stability}

Using the measure of stability defined in Section \ref{sec:stabledef}, we apply the methodology developed in \cite{StableClassification} for building stable neural networks. At a high level, this corresponds to constructing a model that is robust to the specific subset of data used to train it. One way to think about this is to view the training data set as a sample from the true data distribution and then require the model to be robust to the specific sample. Considering the partition of the data into training/validation sets as a sampling mechanism from this true data distribution (each split choice gives a different training set), we desire to build models that are robust to every partition.

To achieve this, we first associate each observation $(\bx_n,y_n)$ with a binary variable $z_n$, $n\in [N]$ that indicates whether or not  $(\bx_n,y_n)$  is part of the training set. We then choose the network's parameters as to minimize the worst-case loss over all possible allocations of these $z_n$'s, resulting in a model that is explicitly built to do well not just over one training set, but over all possible training sets. We start from the same minimization problem introduced in Section \ref{subsec:hdl-network}, i.e.,
$$\min_{\bbbw} \frac{1}{N}\sum_{n=1}^N \mathcal{L}(y_n, \bz^L(\bbbw, \bx_n)).$$
To obtain network stability we require the model to be robust to every training set of fixed size $a$, which results in the following optimization problem: 
\begin{equation}
\begin{aligned}
\label{eq:rob0}
\min_{\bbbw}& \max_{z \in {\cal Z}}\frac{1}{a}\sum_{n=1}^N z_n\mathcal{L}(y_n, \bz^L(\bbbw, \bx_n)), \\
\quad \text{where} \
&{\cal Z}= \left\{z: ~\sum_{n=1}^N z_n = a, \quad  z_n \in \{0,1\},~ n\in[N]\right\}.
\end{aligned}
\end{equation}
The value of $a$ indicates the desired proportion between the size of the training and validation sets. For example, by setting $a = 0.7N$ we recover the typical 70/30 training/validation split. Since the inner maximization problem is linear in $z$, the problem is equivalent to optimizing over the convex hull of ${\cal Z}$. This implies that the binary constraints on $z_n$ can be relaxed to $0\leq z_n\leq 1$, and the inner maximization problem becomes linear in the variables $z_n$. Computing its dual problem we obtain that the value of the inner maximization problem is equivalent to:
$$ \min_{\theta,u_n} ~ \theta + \frac{1}{a}\sum_{n=1}^N u_n \quad \text{subject to} \quad \theta + u_n \geq \mathcal{L}(y_n, \bz^L(\bbbw, \bx_n)), \quad u_n \geq 0,~ n\in[N].$$
Therefore, the stability problem becomes
$$
 \min_{\substack{\bbbw, \\ \theta,u_n\in\mathbb{R}}}  \theta + \frac{1}{a}\sum_{n=1}^N u_n \quad \text{subject to} \quad \theta + u_n \geq \mathcal{L}(y_n, \bz^L(\bbbw, \bx_n)), \:\: u_n \geq 0,~ n\in[N].
$$
Note that the variables $u_n$ can be solved in closed form as $u_n = [\mathcal{L}(y_n, \bz^L(\bbbw, \bx_n)) - \theta]^+$. The final minimization problem with stability then becomes:
$$
 \min_{\bbbw, \theta} \enspace \theta + \frac{1}{a}\sum_{n=1}^N \left[\mathcal{L}(y_n, \bz^L(\bbbw, \bx_n)) - \theta\right]^+,
$$
which is now an unconstrained problem that can be solved with standard gradient descent optimization algorithms.

\section{Experiments}
\label{sec_comp_exp_regr}
This section presents extensive computational experiments comparing the nominal DL approach (abbreviated DL) with 7 other models resulting from our holistic methodology. We showcase the merit of our HDL framework and investigate the influence of each studied component -- robustness, sparsity, and stability -- on the overall performance across 4 evaluation metrics:
\begin{itemize}
    \item \textit{Natural accuracy}: Average accuracy on the testing set across the 10 different train-validation splits with respect to the original input data.
    \item \textit{Adversarial robustness}: Average adversarial accuracy on the testing set across the 10 different train-validation splits with respect to adversarial attacks resulting from perturbations of the original input data. We consider only attacks bounded in the $L_\infty$ norm by some radius $\rho$ using Projected Gradient Descent as in \cite{madry}.
    \item \textit{Stability}: Worst accuracy on the testing set across the 10 different train-validation splits with respect to the original input data.
    \item \textit{Sparsity}: Percentage of network parameters with value $0$.
\end{itemize} 
The exact optimization problem solved for each model results from combinations of the loss functions described in the previous section, and the specific formulations can be found in Table \ref{table:loss_functions} above. 

\begin{table}[t]
\centering
\begin{tabular}{ |P{1cm}|P{9.95cm}| }
\hline
{\centering \textbf{Method} }& {\centering \textbf{Optimization Problem}}  \\
\hline
DL &  $\min_{\bbbw} \: \frac{1}{N}\sum_{n = 1}^N\log\left(\sum_{k} e^{(\Delta \boldsymbol{e}_k^{y_n})^\top\bz^L(\bbbw, \bx_n)}\right)$ \\ 
\hline
Robust & $\min_{\bbbw}\: \frac{1}{N}\sum_{n = 1}^N\log\left(\sum_{k} e^{ (\Delta \boldsymbol{e}_k^{y_n})^\top\bz^L(\bbbw, \bx ) + \rho \|\nabla_{\bx} (\Delta \boldsymbol{e}_k^{y_n})^\top\bz^L(\bbbw, \bx)\|_1}\right)$ \\ 
\hline
Stable & $\min_{\bbbw, \theta}\: \theta + \frac{1}{a}\sum_{n=1}^N [\log\left(\sum_{k} e^{(\Delta \boldsymbol{e}_k^{y_n})^\top\bz^L(\bbbw, \bx_n)}\right) - \theta]^+$ \\ 
\hline
Sparse  & $\min_{\bbbw, \bm{s}} \: \lambda \sum_{j=1}^{\vert\bbbw\vert} \sigma(\beta s_j) + \frac{1}{N}\sum_{n = 1}^N\log\left(\sum_{k} e^{(\Delta \boldsymbol{e}_k^{y_n})^\top\bz^L(\bbbw\odot \sigma(\beta \bm{s}), \bx_n)}\right)$\\
\hline
 Robust & \\
  $+$ & \makecell{$\min_{\bbbw, \bm{s}}\: \lambda \sum_{j=1}^{\vert\bbbw\vert} \sigma(\beta s_j) + \frac{1}{N} \times$ \\ $\sum_{n = 1}^N\!\log\!\left(\!\sum_{k} e^{ (\Delta \boldsymbol{e}_k^{y_n})^\top\!\bz^L(\bbbw\odot \sigma(\beta \bm{s}), \bx ) + \rho \|\nabla_{\bx} (\Delta \boldsymbol{e}_k^{y_n})^\top\!\bz^L(\bbbw, \bx)\|_1}\!\right)$}\\
Sparse & \\
\hline
 Stable  & \\
 + & $\min_{\bbbw, \theta,  \bm{s}}\: \lambda \sum_{j=1}^{\vert\bbbw\vert} \sigma(\beta s_j) + \theta + \frac{1}{a}\sum_{n=1}^N [\log\left(\sum_{k} e^{(\Delta \boldsymbol{e}_k^{y_n})^\top\bz^L(\bbbw\odot \sigma(\beta \bm{s}), \bx_n)}\right) - \theta]^+$\\
Sparse & \\
\hline 
Stable  &\\
+&$ \min_{\bbbw, \theta}\: \theta + \frac{1}{a}\sum_{n=1}^N [\log\left(\sum_{k} e^{ (\Delta \boldsymbol{e}_k^{y_n})^\top\bz^L(\bbbw, \bx ) + \rho \|\nabla_{\bx} (\Delta \boldsymbol{e}_k^{y_n})^\top\bz^L(\bbbw, \bx)\|_1}\right) - \theta]^+$ \\
Robust& \\
\hline 
HDL & \makecell{$ \min_{\bbbw, \theta,  \bm{s}} \lambda \sum_{j=1}^{\vert\bbbw\vert} \sigma(\beta s_j)  + \theta +\frac{1}{a}\times$\\ $\sum_{n=1}^N [\log\sum_{k}\! e^{ (\Delta \boldsymbol{e}_k^{y_n})^\top\bz^L(\bbbw\odot \sigma(\beta \bm{s}), \bx ) + \rho \|\nabla_{\bx} (\Delta \boldsymbol{e}_k^{y_n})^\top\!\bz^L(\bbbw, \bx)\|_1}\! -  \theta]^+ $} \\
\hline
\end{tabular}
\caption{Loss functions used for DL and all methods in the HDL framework.} \label{table:loss_functions}
\end{table}

\paragraph*{Data}  We computed experiments on classification tasks with 45 UCI data sets from the UCI Machine Learning Repository \citep{uci}. These data sets give various problem sizes and difficulties to form a representative sample of real-world tabular problems, with the largest data set having 245,056 observations and the highest number of features being 856. We also benchmarked our methodologies on three image data sets: MNIST, Fashion-MNIST, and CIFAR10.

\paragraph*{Implementation} Our code is written in Python 3.8 \citep{python}. Neural networks are coded using Tensorflow v1 \citep{tensorflow}. We trained each model on a system equipped with an Intel Xeon Gold 6248 processor, which included 4 CPU cores and one Nvidia Volta V100 GPU.

\paragraph*{Training Methodology} For each data set, we used 20\% of the data to obtain a fixed test set, and we applied 80\%/20\% train-validation splits to the remaining data points. Given a choice of model and evaluation metric, we selected the hyperparameters that led to the best average performance in the validation set for the metric in question. We then reported the average performance of the chosen parameter configuration on the test set with respect to the given metric. For all evaluation metrics, the average performance is computed over 10 training-validation splits that are random but identical for all experiments for a fair comparison. 

\paragraph*{Neural network architectures} For our experiments on UCI data sets, we used a feedforward neural network architecture with 2 hidden layers, each with 128 neurons and ReLU activations. For our experiments on the image data sets, we used a convolutional neural network with the AlexNet architecture \citep{alexnet}. We used the Glorot uniform initialization \citep{glorot} for the network weights $\bbbw$ and $\bold{0}$ as initialization for the sparsity variable $\bm{s}_0$.

\paragraph*{Hyperparameter search} 

For each model, we cross-validated the values of the following hyperparameters: 
\begin{itemize}
\item Adam learning rate: \{$1e^{-2}, 1e^{-3}\}$ for UCI data sets, $\{1e^{-3},1e^{-4}\}$ for image data sets.
\item Number of epochs: 150 for UCI data sets, 50 for vision data sets.
\item Batch Size: 32 for UCI data sets, 64 for image data sets.
\item Robustness radius $\rho$ : $\{1e^{-1}, ~1e^{-2}, ~1e^{-3}, ~1e^{-4}, ~1e^{-5}\}$.
\item Sparsity regularization parameter $\lambda$: $\{1e^{-6}, ~1e^{-8}, ~1e^{-10}\}$.
\item Sparsity temperature parameter $\bar{\beta}:$ $\{200, 1000\}$.
\item Stability parameter $a$: \{0.7, 0.8, 0.9, 1\}.
\end{itemize}

\subsection{UCI Data sets}\label{subsection:uci_datasets}
\begin{figure}[H]
     \centering
     \begin{subfigure}[b]{0.49\textwidth}
         \centering
         \includegraphics[width=\textwidth]{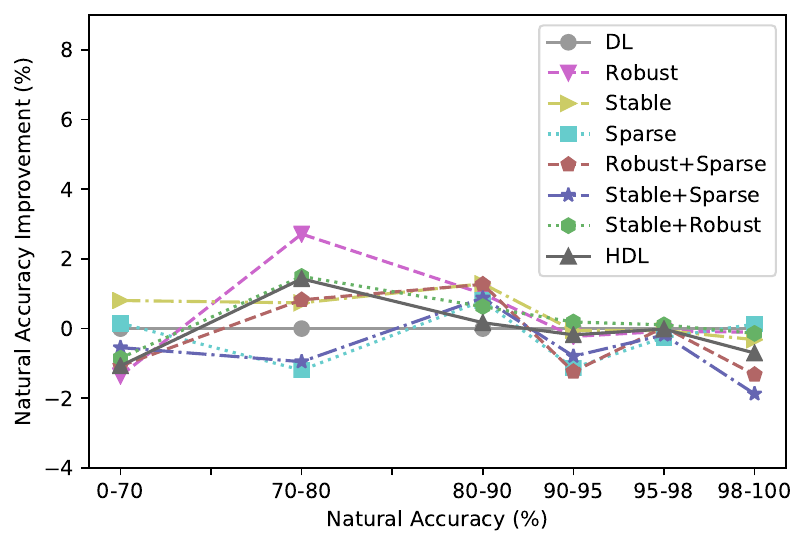}
         \caption{Natural accuracy.}
         \label{fig:uci_acc}
     \end{subfigure}
     \hfill
     \begin{subfigure}[b]{0.49\textwidth}
         \centering
         \includegraphics[width=\textwidth]{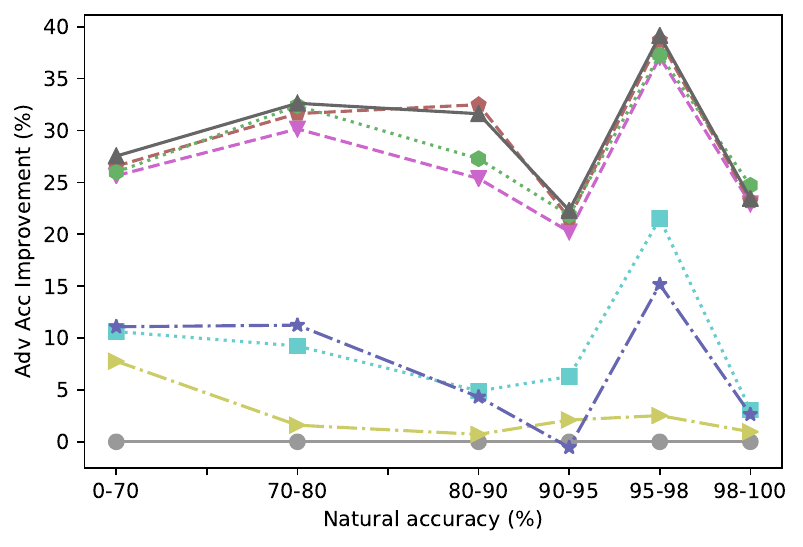}
         \caption{Adversarial accuracy with $\rho=0.1$.}
         \label{fig:uci_adv_acc}
     \end{subfigure}
     \hfill
     \begin{subfigure}[b]{0.49\textwidth}
         \centering
         \includegraphics[width=\textwidth]{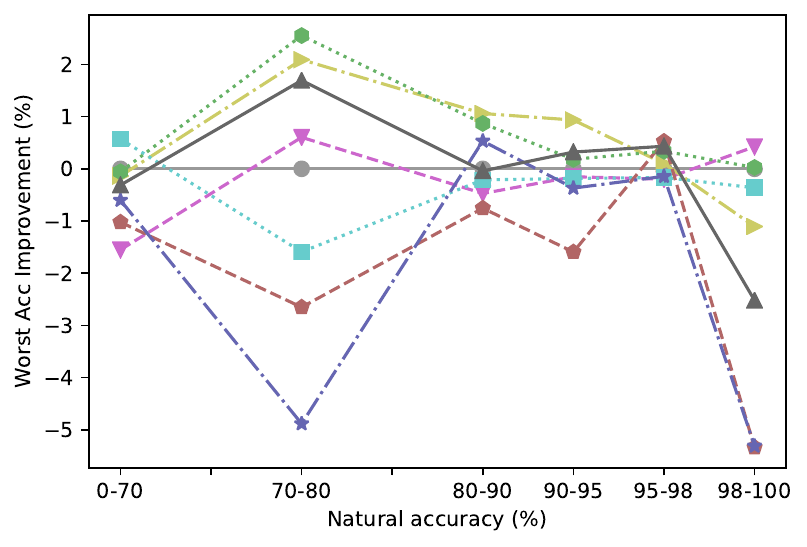}
         \caption{Stability.}
         \label{fig:uci_stability}
     \end{subfigure}
     \hfill
     \begin{subfigure}[b]{0.49\textwidth}
         \centering
         \includegraphics[width=\textwidth]{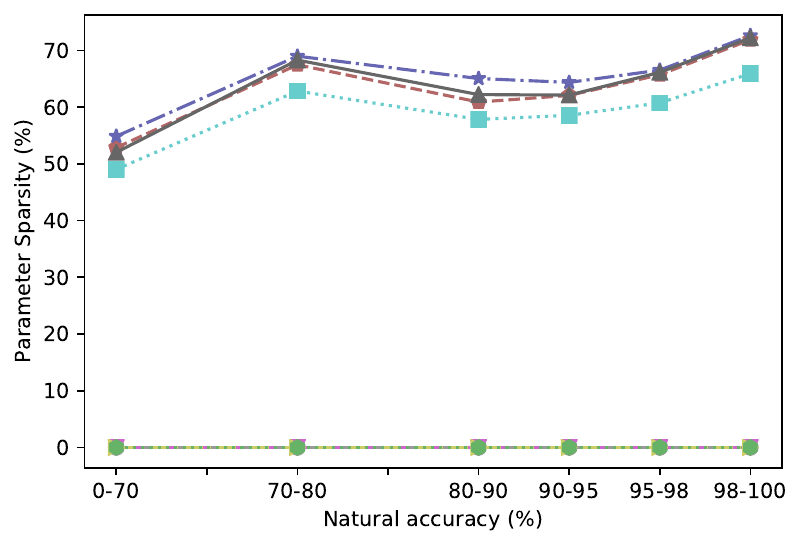}
         \caption{Sparsity.}
         \label{fig:uci_sparsity}
     \end{subfigure}
    \hfill
        \caption{Evaluation of the different methods depending on the natural accuracy of the nominal DL approach on the UCI data sets.}
        \label{fig:results_uci}
\end{figure}

We split the 45 UCI data sets into 6 roughly even-sized groups based on their difficulty level. Specifically, we consider the ranges $0\%$-$70\%$, $70\%$-$80\%$, $80\%$-$90\%$, $90\%$-$95\%$, $95\%$-$98\%$ and $98\%$-$100\%$ of natural accuracy achieved by the nominal DL approach. We first investigate the performance of the HDL framework with respect to a single evaluation metric. In Figure \ref{fig:results_uci}, we evaluate all methods in terms of natural accuracy, adversarial accuracy with $\rho = 0.1$, stability, and sparsity.

Figures \ref{fig:uci_acc} and \ref{fig:uci_stability} show that those data sets for which the nominal approach achieves accuracy in the $70\%$-$90\%$ range are the ones that benefit the most from the HDL framework (especially the Robust, Stable, and Stable+Robust models) when the evaluation metric corresponds to natural accuracy or stability. For the data sets with natural accuracy above $90\%$, none of the models significantly improve over the natural accuracy or stability achieved by the nominal DL model. However, for data sets in the $98-100\%$ range sparsity slightly improves accuracy and robustness slightly helps for stability. 

\begin{figure}[H]
    \centering
    \includegraphics[width=0.65\textwidth]{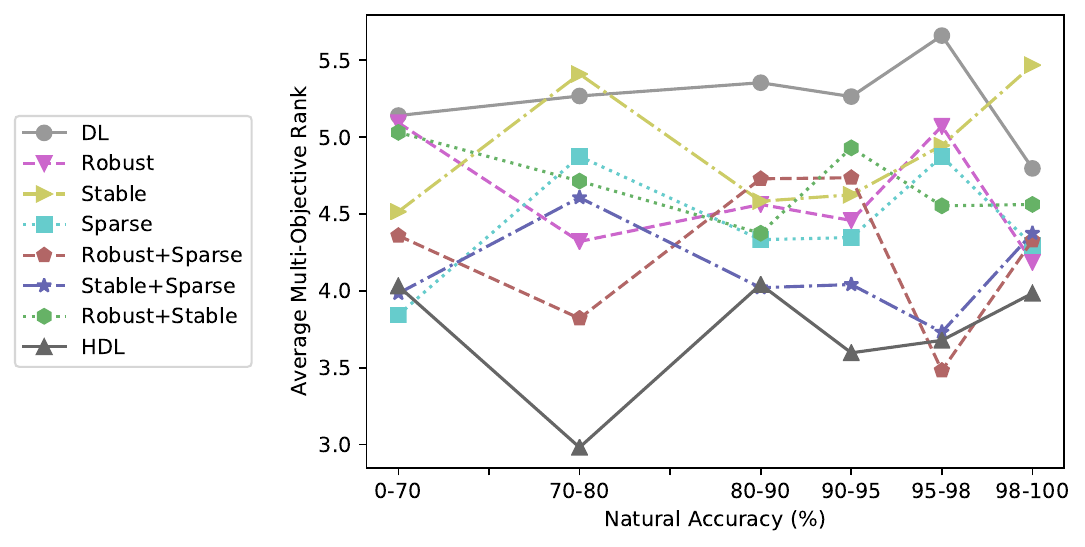}
    \caption{Average multi-objective rank.}
    \label{fig:uci_multi-objective}
\end{figure}

Figure \ref{fig:uci_adv_acc} shows the adversarial robustness achieved with perturbation parameter $\rho = 0.1$. We see a substantial adversarial robustness improvement in all methods that included the robust component. Moreover, combining robustness with stability and/or sparsity leads to higher adversarial accuracy than that achieved with robustness alone.  In terms of parameter sparsity, Figure \ref{fig:uci_sparsity} shows that all models with imposed sparsity (Sparse, Stable$+$Sparse, Robust+Sparse, and HDL) have a much lower percentage of nonzero parameters compared to the models without it. And importantly, both robustness and stability help achieve sparser models when combined with sparsity.
Since we are also interested in models that are simultaneously accurate, sparse, robust, and stable, we consider a multi-objective metric using the rank of each method (ranks start at 1, with lower ranks corresponding to better performance). For each method, we use the natural accuracy, adversarial accuracy, stability, and sparsity achieved in the validation set respectively to rank all its hyperparameter configurations 4 times. Then for each hyperparameter configuration, we compute the average rank across the 4 metrics and select the configuration that leads to the method's highest average rank. Finally, we rank the 8 selected models (for the 8 different methods) with respect to each evaluation metric on the testing set to obtain their out-of-sample average rank. As shown in Figure \ref{fig:uci_multi-objective}, all 7 models from the HDL framework outperform the nominal DL approach with respect to this holistic metric. Moreover, the HDL model typically achieves the best results across data set complexities.
\vspace{-0.3cm}
\vspace{-0.7cm}
\subsection{Image Data Sets}
In this section, we evaluate all methods using the MNIST, Fashion-MNIST, and CIFAR10 data sets. For each method, we select the parameters based on the multi-objective metric utilized for the UCI data sets in the validation set and report the performance across metrics. In Tables \ref{tab:results_fashion_mnist_weighted} and \ref{tab:results_mnist_weighted}, we see that for MNIST and Fashion-MNIST, the HDL model outperforms the DL model for all objectives. In particular, HDL achieves higher accuracy using only around 70\% of the parameters. The results for the CIFAR10 data set (Table \ref{tab:results_cifar_weighted}) are a bit different since adding sparsity slightly hurts natural accuracy. However, the accuracy achieved by the HDL model is comparable to those achieved by the non-sparse models and the number of parameters is reduced by 47\%.

\begin{table}[hbt!]
\fontsize{9}{9}\selectfont
\resizebox{\textwidth}{!}{
\begin{tabular}
{|c|c|c|c|c|c|c|c|c|}
\hline
     \textbf{Method} & \textbf{DL} & \textbf{Rob.} & \textbf{Stab.} & \textbf{Sparse} & \textbf{Rob. +} & \textbf{Stab. +} & \textbf{Stab. +} & \textbf{HDL} \\
      &  &  &  & & \textbf{Sparse} & \textbf{Sparse} & \textbf{Rob.} & \\\hline
     Avg. Accuracy & $91.8$ & $92.0$& $91.9$& $91.4$& {$\boldsymbol{92.1}$}& $91.4$& {92.0}& $\boldsymbol{92.1}$\\
     Adv. Acc. $\rho=0.01$ & $78.7$ & $81.1$ & $78.3$ & $80.8$ & $86.9$ & $80.2$ & $86.8$ & $\boldsymbol{87.1}$\\
     Stability & $91.5$ & $91.7$ & $\boldsymbol{91.8}$ & $91.3$& $91.9$ & $91.2$ & ${91.7}$ & $\boldsymbol{91.8}$ \\
     Sparsity & $0$ & $0$ & $0$& $36.2$& $26.6$& \textbf{48.4}& $0$& $26.8$\\\hline 
\end{tabular}}
\caption{Results for the Fashion-MNIST data set. For each method, the parameters with the highest average rank in the validation set were chosen.}
\label{tab:results_fashion_mnist_weighted}
\end{table}
\begin{table}[htp]
\fontsize{9}{9}\selectfont
\resizebox{\textwidth}{!}{
\begin{tabular}{|c|c|c|c|c|c|c|c|c|}
\hline
     \textbf{Method} & \textbf{DL} & \textbf{Rob.} & \textbf{Stab.} & \textbf{Sparse} & \textbf{Rob. +} & \textbf{Stab. +} & \textbf{Stab. +} & \textbf{HDL} \\
      &  &  &  & & \textbf{Sparse} & \textbf{Sparse} & \textbf{Rob.} & \\\hline
     Avg. Accuracy & $99.2$ & $\boldsymbol{99.3}$ & $99.2$ & $99.1$ & $99.2$ & $99.2$ & $\boldsymbol{99.3}$ & $\boldsymbol{99.3}$\\
     Adv. Acc. $\rho = 0.1$ & $49.6$ & $78.4$ & $51.5$ & $42.6$ & $74.7$ & $27.7$ & $\boldsymbol{79.5}$ & $76.0$\\
     Stability & $99.1$ & $99.2$ & $99.2$ & $99.1$ & $99.1$ & $99.0$ & $\boldsymbol{99.3}$ & $99.2$ \\
     Sparsity &$0$ & $0$ & $0$ & $16.1$ & $27.9$ & $\boldsymbol{31.7}$ & $0$ & $28.0$\\\hline 
\end{tabular}}
\caption{Results for the MNIST data set. For each method, the parameters with the highest average rank in the validation set were chosen.}
\label{tab:results_mnist_weighted}
\end{table}
\vspace{-0.5cm}
\begin{table}[htp]
\fontsize{9}{9}\selectfont
\resizebox{\textwidth}{!}{
\begin{tabular}{|c|c|c|c|c|c|c|c|c|}
\hline
     \textbf{Method} & \textbf{DL} & \textbf{Rob.} & \textbf{Stab.} & \textbf{Sparse} & \textbf{Rob. +} & \textbf{Stab. +} & \textbf{Stab. +} & \textbf{HDL} \\
      &  &  &  & & \textbf{Sparse} & \textbf{Sparse} & \textbf{Rob.} & \\\hline

     Avg. Accuracy & $\boldsymbol{70.1}$ & $\boldsymbol{70.1}$ & $\boldsymbol{70.1}$ & $69.8$ & $69.2$ & $69.3$ & $69.8$ & $69.3$\\
     Adv. Acc. $\rho = 0.01$ & $26.7$ & $27.1$ & $26.6$ & $27.3$ & $27.4$ & $27.7$ & $29.1$ & $\boldsymbol{30.6}$\\
     Stability & $69.7$ & $69.7$ & $\boldsymbol{69.8}$ & $69.3$ & $68.9$ & $68.5$ & $69.4$ & $68.8$  \\
     Sparsity &$0$ & $0$ & $0$ & $28.7$ & $47.2$ & $\boldsymbol{47.8}$ & $0$ & $\boldsymbol{47.8}$\\\hline 
\end{tabular}}
\caption{Results for the CIFAR10 data set. For each method, the parameters with the highest average rank in the validation set were chosen.}
\label{tab:results_cifar_weighted}
\end{table}
\vspace{-0.5cm}
\FloatBarrier
\subsection{Computational Times}

Since modifying the loss function often affects the training computational time, we quantify the slowdown effect for all the methods in the HDL framework. Specifically, for each of the 45 UCI data sets as well as the 3 image data sets introduced in the previous section, we calculate how many times slower each method is when compared to the nominal DL approach in terms of batches per second as well as number of iterations needed. The average slowdown factors are shown in Table \ref{tab:slowing_factors}. We observe that robustness and sparsity both decrease the number of batches per second by approximately a factor of $3$, while stability preserves the same speed as the DL approach. In addition, since we used 5 training rounds for the methods incorporating sparsity, they require 5 times as many training iterations as the other methods. On average, the HDL method is only 16 times slower, and methods that don't optimize for sparsity only increase the computational time by less than 3 times. 

\vspace{0.5cm}
\begin{table}[ht]
\fontsize{9}{9}\selectfont
\resizebox{\textwidth}{!}{
\begin{tabular}{|c|c|c|c|}
\hline
     \textbf{Method} & \textbf{Batches/sec} & \textbf{ No. Iterations} &  \textbf{Total Slowdown} \\ 
      & \textbf{Slowdown Factor} & \textbf{Increase Factor} &  \textbf{factor} \\ 
     \hline
Robust       &                  2.7 &                        1 &                       2.7 \\
Stable     &                   1.0 &                        1 &                       1.0 \\
Sparse     &                  1.2 &                        5 &                       5.9 \\
Robust+Sparse     &                  3.2 &                        5 &                      16.1 \\
Stable+Sparse  &                  1.1 &                        5 &                       5.5 \\
Stable+Robust &                  2.7&                        1 &                       2.7 \\
HDL       &                  3.2 &                        5 &                      16.2 \\
\hline
\end{tabular}}
\caption{Average slowdown factors of computational time with respect to the nominal DL method.}
\label{tab:slowing_factors}
\end{table}
\FloatBarrier

\subsection{SHAP Values}
\begin{figure}[H]
     \centering
     \begin{subfigure}[b]{0.49\textwidth}
         \centering
         \includegraphics[width=\textwidth]{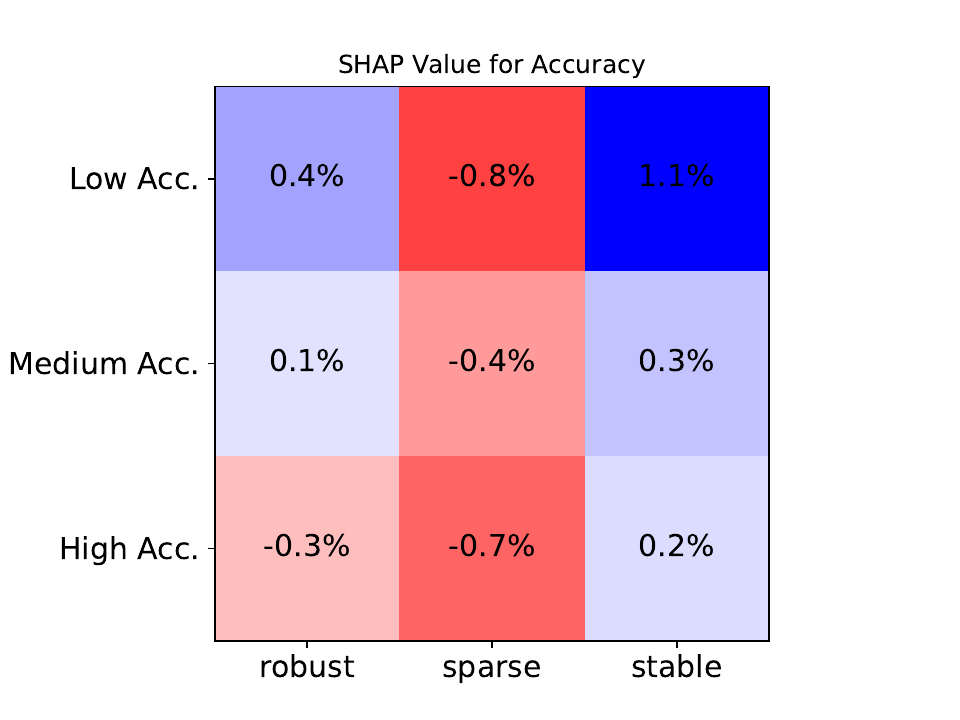}
         \caption{Accuracy.}
         \label{fig:shap_uci_acc}
     \end{subfigure}
     \hfill
     \begin{subfigure}[b]{0.49\textwidth}
         \centering
         \includegraphics[width=\textwidth]{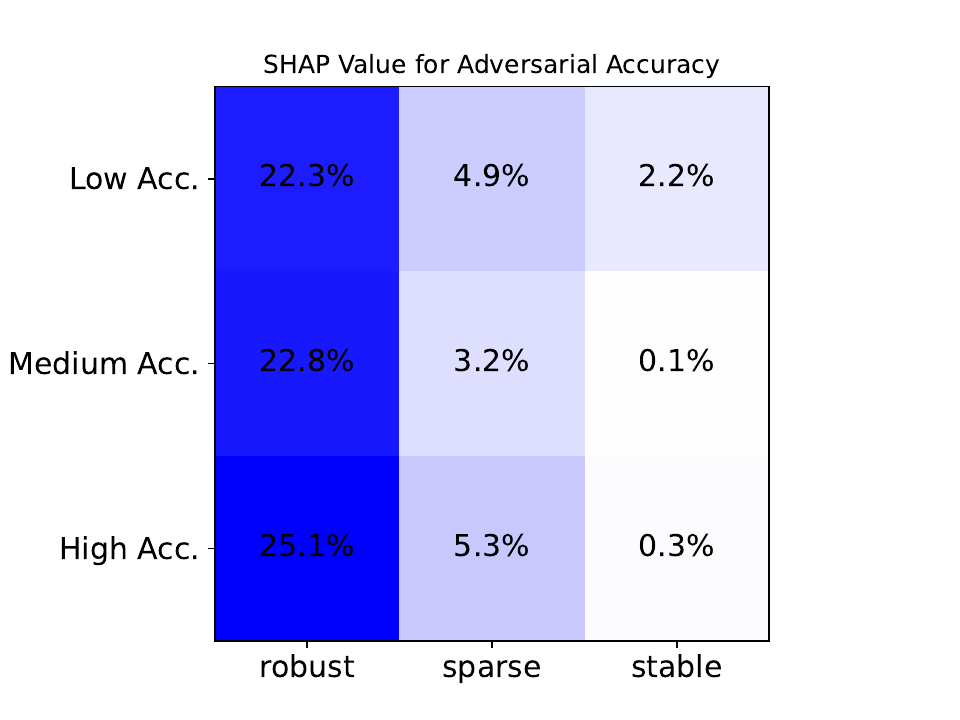}
         \caption{Adversarial accuracy with $\rho=0.1$.}
         \label{fig:shap_uci_adv_acc}
     \end{subfigure}
     \hfill
     \begin{subfigure}[b]{0.49\textwidth}
         \centering
         \includegraphics[width=\textwidth]{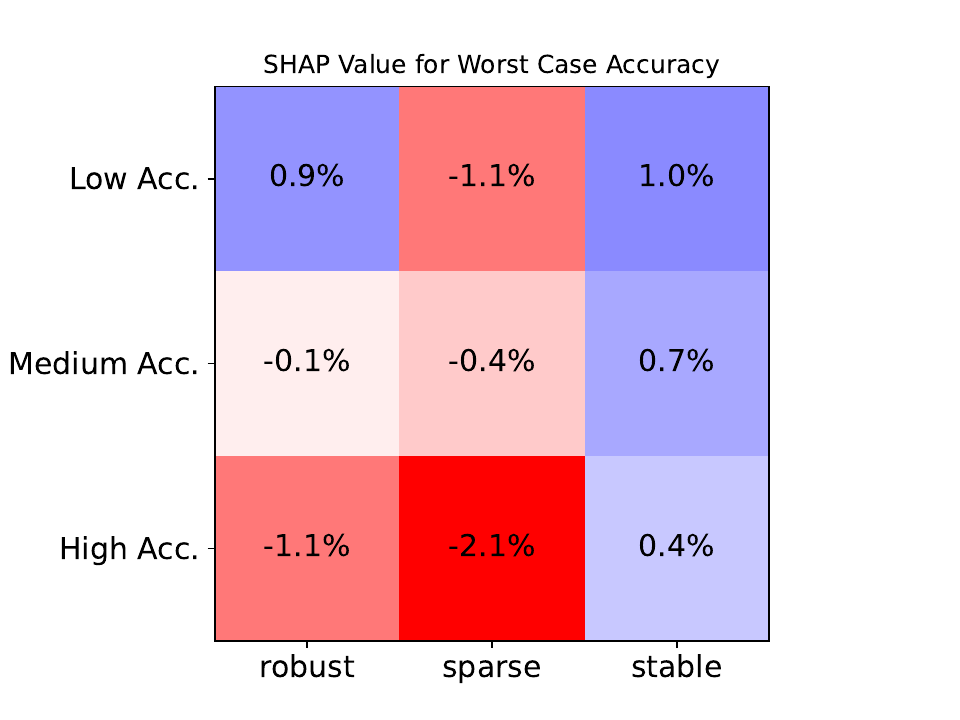}
         \caption{Stability.}
         \label{fig:shap_uci_stability}
     \end{subfigure}
     \hfill
     \begin{subfigure}[b]{0.49\textwidth}
         \centering
         \includegraphics[width=\textwidth]{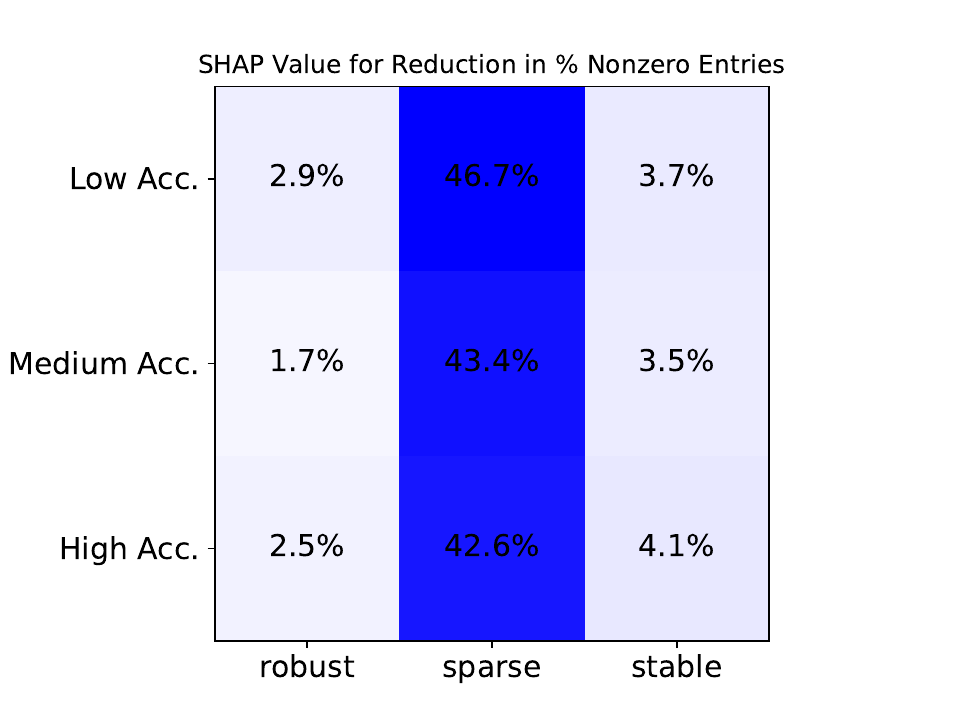}
         \caption{Sparsity.}
         \label{fig:shap_uci_sparsity}
     \end{subfigure}
        \caption{SHAP values on various metrics across different UCI data set categories. Blue/red indicates that the feature has a positive/negative SHAP value on a specific category of UCI data set.}
        \label{fig:shap_uci}
\end{figure}

To gain a deeper understanding of the interplay between individual loss components (robustness, stability, sparsity) and the metrics we measure, we employ the SHAP values method \citep{lundberg2017unified}. We compute the SHAP values for each UCI data set and average the results over three data set categories: Low Accuracy ($<80\%$), Medium Accuracy ($80\%$-$95\%$), and High Accuracy ($>95\%$), with 15 data sets each. The results are shown in Figure \ref{fig:shap_uci}.
\vspace{-0.3cm}

\vspace{-0.5cm}
Our findings confirm that robustness, stability, and sparsity techniques improve the corresponding metrics across all data set categories. More intriguingly, these techniques also positively impact metrics beyond their intended purposes. For example, sparsity and stability enhance adversarial accuracy, while robustness and stability yield sparser networks. This indicates that combining techniques does not necessarily result in any adverse effects and that it is feasible to attain networks with good performance across all metrics. Additionally, the benefits of these techniques are more pronounced in data sets with low initial accuracy, particularly for the accuracy and stability metrics. Lastly, we observe that sparsity generally hurts accuracy and stability, although this highly varies across data sets, as observed in Section \ref{subsection:uci_datasets}.

\subsection{Prescriptive Approach}

In this section, we develop a prescriptive approach that allows users to choose a training loss function based on the specific objective they wish to maximize, which can be a single evaluation metric or a weighted combination of several metrics. Depending on the data set characteristics and the performance scores of the nominal DL model, we propose a tree-based recommendation model to suggest the most suitable HDL loss function for optimal results with respect to the desired objective.

We train our models using an Optimal Policy Tree (OPT) algorithm \citep{amram}, which uses observational data of the form ${(\mathbf{x}_i, y_i, z_i)}$. In our case, each observation (i.e., data set) $i$ encompasses:
\begin{itemize}
    \item Dataset features $\mathbf{x}_i \in \mathbb{R}^{8}$: number of features, number of target classes, nominal DL accuracy, nominal DL adversarial accuracy with $\rho = 0.001$, nominal DL adversarial accuracy with $\rho = 0.01$, nominal DL adversarial accuracy with $\rho = 0.1$, nominal DL stability, nominal DL worst case accuracy.
    \item Prescriptions $z_i\in {1,\ldots,8}$: DL, Robust, Stable, Sparse, Robust+Sparse, Stable+Sparse, Stable+Robust, HDL.
    \item Outcomes $\bm{y}_i\in \mathbb{R}^8$, which represent the performance improvement of each method compared to the nominal DL model with respect to the metric set by the user.
\end{itemize}
Our prescriptive task is to find the optimal policy that, given the information $\mathbf{x}$ of a data set, prescribes the method $z$ leading to the best metric score $y$. We randomly split the 45 UCI data sets into a training set (40 data sets) and a test set (5 data sets from different difficulty levels). We cross-validated the optimal tree depth and complexity using the training set.

\begin{figure}
\begin{minipage}{0.42\textwidth}
  \centering
  \vspace{3.55em}
  \includegraphics[width=\linewidth]{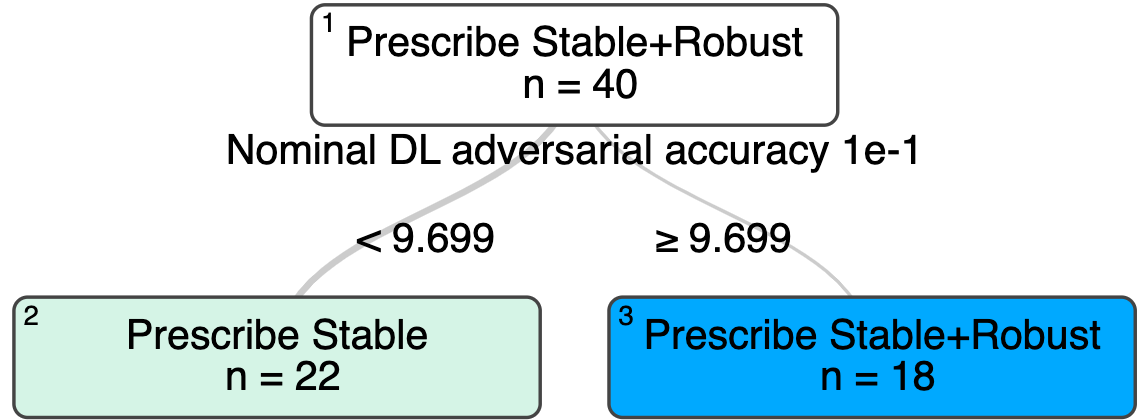}
  \caption{Optimal policy tree for maximizing natural accuracy.}
  \label{fig:natural}
\end{minipage}
\hspace{0.02\textwidth}
\begin{minipage}{0.54\textwidth}
  \centering
  \includegraphics[width=\linewidth]{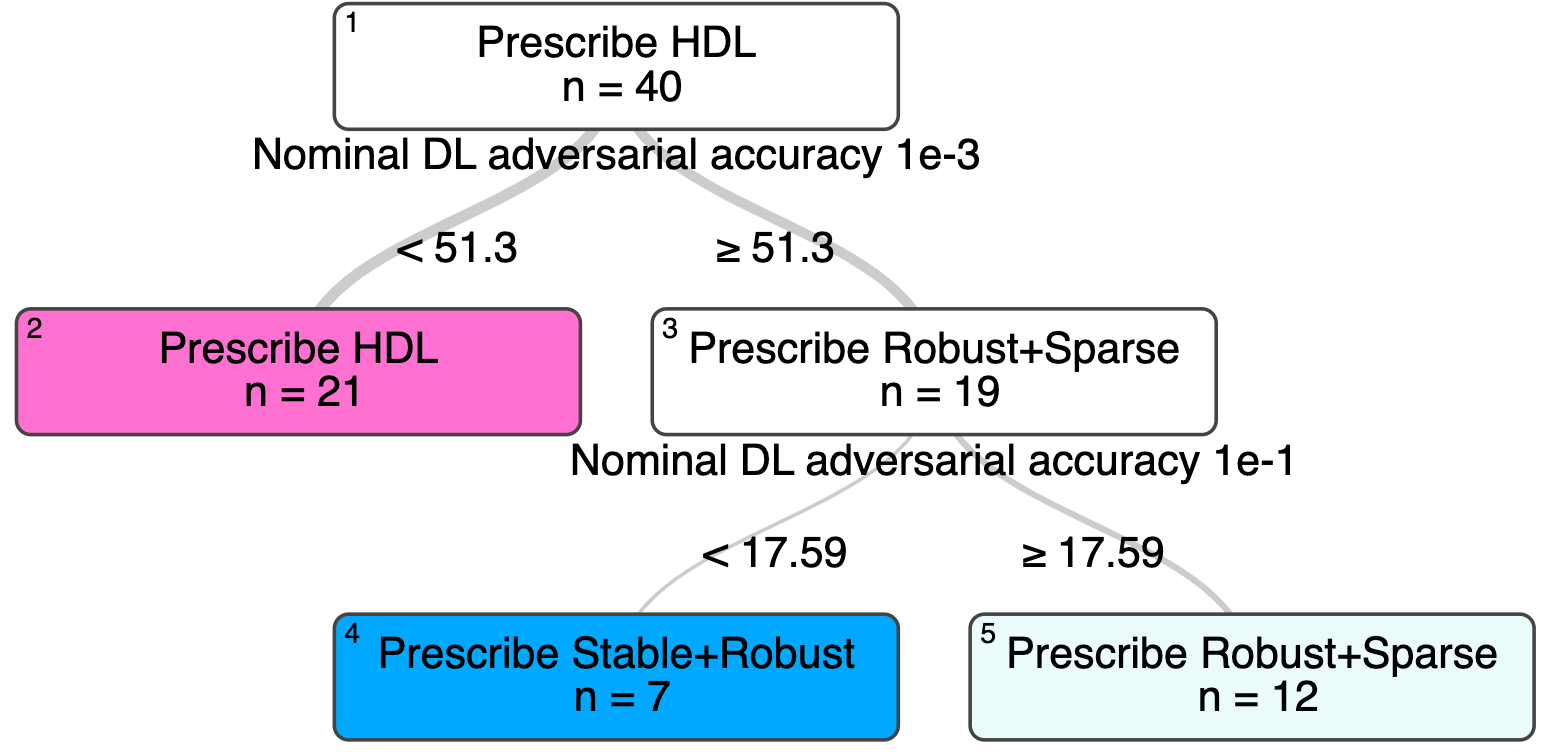}
  \caption{Optimal policy tree for maximizing robustness ($\rho = 0.1$).}
  \label{fig:robust}
\end{minipage}
\end{figure}

Figures \ref{fig:natural} and \ref{fig:robust} represent the OPTs obtained for maximizing two different objectives: natural accuracy and adversarial accuracy. The tree in Figure \ref{fig:natural} highlights that the Stable and Stable+Robust methods are the best suited to obtain high natural accuracy, with the former being preferred when the nominal DL approach has very low adversarial accuracy ($\rho = 0.1$). To maximize robustness, the tree in Figure \ref{fig:robust} prescribes HDL, Stable+Robust, or Robust+Sparse depending on the adversarial accuracy achieved by the nominal DL method. 

In addition, we obtained single-leaf trees when maximizing the stability and sparsity objectives. The recommended methods are Stable+Robust for optimizing stability and Stable+Sparse for maximizing sparsity. Lastly, HDL was always the prescribed method when the desired objective was the equally weighted average of all 4 previous metrics.

Finally, Table \ref{tab:optequal} reports the out-of-sample performance of these prescription trees on the 5 UCI data sets from the test set (cnae-9, hill-valley, libras-movement, magic-gamma, and thyroid-ann). We emphasize that the performance of the prescribed methods is higher than that of the nominal DL approach across all objectives and data sets, and it often matches the performance of the best method. 

\begin{table}[!ht]
    \centering
    \begin{adjustbox}{width=1\textwidth}
    
    \begin{tabular}{|c|c|ccccc|}
    \hline
        \multirow{2}{*}{\textbf{Objective}}& \multirow{2}{*}{\textbf{Method}}& \multicolumn{5}{c|}{\textbf{Test Data Sets Objective Value}} \\
         & & cnae-9	& hill-& libras-&magic-&thyroid-\\
         & & & valley& move&gamma&ann\\
        \hline
        \multirow{3}{*}{Nat. Acc.} & DL & 93.70 & 47.21& 79.44& 87.11& 98.86\\
        & Prescribed & 94.07 &53.61 & 80.00& 87.28& 99.05\\
        & Optimal &94.07 & 53.61 &  82.50& 87.28 &99.05\\
    \hline
    \multirow{3}{*}{Robustness ($\rho = 0.1$)} & DL &0.00 &7.54 &0.00 &15.07 &48.42 \\
        & Prescribed &3.80 &36.39 &2.50 &64.59 &91.79\\
        & Optimal & 3.80& 40.16&4.72 &64.59 &91.79\\
    \hline
    \multirow{3}{*}{Stability} & DL &91.20 &43.44 &75.00 &86.65 &98.28 \\
        & Prescribed & 93.06& 45.08&81.94 &87.01 &98.54\\
        & Optimal & 93.06&49.18 &81.94 &87.01 &98.81\\
    \hline
    \multirow{3}{*}{Sparsity} & DL & 0.00 &0.00 &0.00 &0.00 &0.00 \\
        & Prescribed & 73.43&34.89 &71.00 &57.52 &53.22\\
        & Optimal & 73.43&41.94 &71.00 &57.52 &53.22\\
    \hline
   (\text{Nat. Acc.} $+$\text{Robustness} & DL & 46.25& 24.55& 38.61& 47.21 & 61.39 \\
       $+$\text{Stability} & Prescribed & 57.75& 39.35 & 52.76 & 58.06 & 73.03\\
       $+$\text{Sparsity})/4& Optimal & 62.07 & 40.66 & 52.82 & 59.62 & 73.03\\
    \hline
    \end{tabular}
    \end{adjustbox}
    \caption{Performance of prescription trees on the testing set.}\label{tab:optequal}
\end{table}
\vspace{-0.5cm}
\section{Conclusions}\label{sec:conclusions}

This paper presents a unifying methodology to obtain deep learning models that are accurate, robust, stable, and sparse by appropriately modifying the objective function to be minimized. Across multiple computational experiments, we show how these 4 metrics interact and demonstrate that we can often train models that simultaneously improve adversarial accuracy, worst-case accuracy, and parameter sparsity without sacrificing natural accuracy. Finally, we provide prescriptive trees that recommend which method is more appropriate depending on the desired objective to be maximized.

\newpage
\bibliography{hdl}

\begin{thebibliography}{59}
\providecommand{\natexlab}[1]{#1}
\providecommand{\url}[1]{{#1}}
\providecommand{\urlprefix}{URL }
\providecommand{\doi}[1]{\url{https://doi.org/#1}}
\providecommand{\eprint}[2][]{\url{#2}}
 \bibcommenthead

\bibitem[{Abadi et~al(2015)Abadi, Agarwal, and et~al}]{tensorflow}
Abadi M, Agarwal A, et~al (2015) {TensorFlow}: Large-scale machine learning on
  heterogeneous systems. \urlprefix\url{https://www.tensorflow.org/}, software
  available from tensorflow.org

\bibitem[{Aghasi et~al(2020)Aghasi, Abdi, and Romberg}]{aghasi}
Aghasi A, Abdi A, Romberg J (2020) Fast convex pruning of deep neural networks.
  SIAM Journal on Mathematics of Data Science 2(1):158--188

\bibitem[{Amram et~al(2022)Amram, Dunn, and Zhuo}]{amram}
Amram M, Dunn J, Zhuo YD (2022) Optimal policy trees. Machine Learning
  111:2741–2768

\bibitem[{Anderson et~al(2020)Anderson, Huchette, Ma, Tjandraatmadja, and
  Vielma}]{anderson2020strong}
Anderson R, Huchette J, Ma W, et~al (2020) Strong mixed-integer programming
  formulations for trained neural networks. Mathematical Programming pp 1--37

\bibitem[{Athalye et~al(2018)Athalye, Carlini, and
  Wagner}]{athalye2018obfuscated}
Athalye A, Carlini N, Wagner D (2018) Obfuscated gradients give a false sense
  of security: Circumventing defenses to adversarial examples.
  \eprint{1802.00420}

\bibitem[{Bellec et~al(2017)Bellec, Kappel, Maass, and Legenstein}]{bellec}
Bellec G, Kappel D, Maass W, et~al (2017) Deep rewiring: Training very sparse
  deep networks. CoRR abs/1711.05136.
  \urlprefix\url{http://arxiv.org/abs/1711.05136},
  {\href{https://arxiv.org/abs/1711.05136}{{https://arxiv.org/abs/arXiv:1711.05136}}}

\bibitem[{Bertsimas and Paskov(2020)}]{StableRegression}
Bertsimas D, Paskov I (2020) Stable regression: On the power of optimization
  over randomization in training regression problems. Journal of Machine
  Learning Research 21(230):1--25

\bibitem[{Bertsimas et~al(2020)Bertsimas, Pauphilet, and Parys}]{sparse-reg}
Bertsimas D, Pauphilet J, Parys BV (2020) {Sparse Regression: Scalable
  Algorithms and Empirical Performance}. Statistical Science 35(4):555 -- 578.
  \doi{10.1214/19-STS701}, \urlprefix\url{https://doi.org/10.1214/19-STS701}

\bibitem[{Bertsimas et~al(2021)Bertsimas, Boix, Carballo, and den
  Hertog}]{DBRobustness}
Bertsimas D, Boix X, Carballo KV, et~al (2021) A robust optimization approach
  to deep learning. CoRR abs/2112.09279.
  \urlprefix\url{https://arxiv.org/abs/2112.09279},
  {\href{https://arxiv.org/abs/2112.09279}{{https://arxiv.org/abs/2112.09279}}}

\bibitem[{Bertsimas et~al(2022)Bertsimas, Dunn, and
  Paskov}]{StableClassification}
Bertsimas D, Dunn J, Paskov I (2022) Stable classification. Journal of Machine
  Learning Research 23(296):1--53

\bibitem[{Carlini and Wagner(2017)}]{adverse_carlini}
Carlini N, Wagner D (2017) Towards evaluating the robustness of neural
  networks. In: 2017 {IEEE} symposium on security and privacy (sp), pp 39--57

\bibitem[{Changpinyo et~al(2017)Changpinyo, Sandler, and
  Zhmoginov}]{changpinyo2017power}
Changpinyo S, Sandler M, Zhmoginov A (2017) The power of sparsity in
  convolutional neural networks. arXiv preprint arXiv:170206257

\bibitem[{Cohen et~al(2019)Cohen, Rosenfeld, and Kolter}]{cohen2019certified}
Cohen J, Rosenfeld E, Kolter Z (2019) Certified adversarial robustness via
  randomized smoothing. In: International Conference on Machine Learning, PMLR,
  pp 1310--1320

\bibitem[{Deng(2012)}]{mnist}
Deng L (2012) The mnist database of handwritten digit images for machine
  learning research. IEEE Signal Processing Magazine 29(6):141--142

\bibitem[{Dua and Graff(2017)}]{uci}
Dua D, Graff C (2017) {UCI} machine learning repository.
  \urlprefix\url{http://archive.ics.uci.edu/ml}

\bibitem[{Dvijotham et~al(2018)Dvijotham, Stanforth, Gowal, Mann, and
  Kohli}]{dvijotham2018dual}
Dvijotham K, Stanforth R, Gowal S, et~al (2018) A dual approach to scalable
  verification of deep networks. In: UAI, p~3

\bibitem[{Gale et~al(2019)Gale, Elsen, and Hooker}]{stateofsparsity}
Gale T, Elsen E, Hooker S (2019) The state of sparsity in deep neural networks.
  CoRR abs/1902.09574. \urlprefix\url{http://arxiv.org/abs/1902.09574},
  {\href{https://arxiv.org/abs/1902.09574}{{https://arxiv.org/abs/arXiv:1902.09574}}}

\bibitem[{Glorot and Bengio(2010)}]{glorot}
Glorot X, Bengio Y (2010) Understanding the difficulty of training deep
  feedforward neural networks. In: Teh YW, Titterington M (eds) Proceedings of
  the Thirteenth International Conference on Artificial Intelligence and
  Statistics, Proceedings of Machine Learning Research, vol~9. PMLR, Chia
  Laguna Resort, Sardinia, Italy, pp 249--256,
  \urlprefix\url{https://proceedings.mlr.press/v9/glorot10a.html}

\bibitem[{Goldwasser et~al(2020)Goldwasser, Kalai, Kalai, and Montasser}]{DR2}
Goldwasser S, Kalai AT, Kalai Y, et~al (2020) Beyond perturbations: Learning
  guarantees with arbitrary adversarial test examples. Advances in Neural
  Information Processing Systems 33:15,859--15,870

\bibitem[{Goodfellow et~al(2014)Goodfellow, Shlens, and
  Szegedy}]{adverse_goodfellow}
Goodfellow IJ, Shlens J, Szegedy C (2014) Explaining and harnessing adversarial
  examples. \eprint{1412.6572}

\bibitem[{Han et~al(2015)Han, Pool, Tran, and Dally}]{han_sparse}
Han S, Pool J, Tran J, et~al (2015) Learning both weights and connections for
  efficient neural network. Advances in neural information processing systems
  28

\bibitem[{Hastie et~al(2001)Hastie, Tibshirani, and Friedman}]{ESL}
Hastie T, Tibshirani R, Friedman J (2001) The Elements of Statistical Learning.
  Springer Series in Statistics, Springer New York Inc., New York, NY, USA

\bibitem[{Hein and Andriushchenko(2017)}]{DBLP:journals/corr/HeinA17}
Hein M, Andriushchenko M (2017) Formal guarantees on the robustness of a
  classifier against adversarial manipulation. Advances in neural information
  processing systems 30

\bibitem[{Hoefler et~al(2021)Hoefler, Alistarh, Ben-Nun, Dryden, and
  Peste}]{sparsityindeep}
Hoefler T, Alistarh D, Ben-Nun T, et~al (2021) Sparsity in deep learning:
  Pruning and growth for efficient inference and training in neural networks.
  The Journal of Machine Learning Research 22(1):10,882--11,005

\bibitem[{Ilyas et~al(2017)Ilyas, Jalal, Asteri, Daskalakis, and
  Dimakis}]{samangouei}
Ilyas A, Jalal A, Asteri E, et~al (2017) The robust manifold defense:
  Adversarial training using generative models. CoRR abs/1712.09196.
  \urlprefix\url{http://arxiv.org/abs/1712.09196},
  {\href{https://arxiv.org/abs/1712.09196}{{https://arxiv.org/abs/arXiv:1712.09196}}}

\bibitem[{Janowsky(1989)}]{janowski}
Janowsky SA (1989) Pruning versus clipping in neural networks. Phys Rev A
  39:6600--6603. \doi{10.1103/PhysRevA.39.6600},
  \urlprefix\url{https://link.aps.org/doi/10.1103/PhysRevA.39.6600}

\bibitem[{Kabilan et~al(2021)Kabilan, Morris, Nguyen, and Nguyen}]{kabilan}
Kabilan VM, Morris B, Nguyen HP, et~al (2021) Vectordefense: Vectorization as a
  defense to adversarial examples. Soft Computing for Biomedical Applications
  and Related Topics pp 19--35

\bibitem[{Krizhevsky et~al(2009)Krizhevsky, Hinton et~al}]{cifar}
Krizhevsky A, Hinton G, et~al (2009) Learning multiple layers of features from
  tiny images. Tech. rep.

\bibitem[{Krizhevsky et~al(2012)Krizhevsky, Sutskever, and Hinton}]{alexnet}
Krizhevsky A, Sutskever I, Hinton GE (2012) Imagenet classification with deep
  convolutional neural networks. In: Proceedings of the 25th International
  Conference on Neural Information Processing Systems - Volume 1. Curran
  Associates Inc., Red Hook, NY, USA, NIPS'12, p 1097–1105

\bibitem[{Krogh and Vedelsby(1994)}]{CV1}
Krogh A, Vedelsby J (1994) Neural network ensembles, cross validation, and
  active learning. Advances in neural information processing systems 7

\bibitem[{Lamb et~al(2018)Lamb, Binas, Goyal, Serdyuk, Subramanian, Mitliagkas,
  and Bengio}]{lamb}
Lamb A, Binas J, Goyal A, et~al (2018) Fortified networks: Improving the
  robustness of deep networks by modeling the manifold of hidden
  representations. \eprint{1804.02485}

\bibitem[{LeCun et~al(1989)LeCun, Denker, and Solla}]{lecun}
LeCun Y, Denker J, Solla S (1989) Optimal brain damage. Advances in neural
  information processing systems 2

\bibitem[{Lecuyer et~al(2019)Lecuyer, Atlidakis, Geambasu, Hsu, and
  Jana}]{lecuyer2019certified}
Lecuyer M, Atlidakis V, Geambasu R, et~al (2019) Certified robustness to
  adversarial examples with differential privacy. In: 2019 IEEE Symposium on
  Security and Privacy (SP), IEEE, pp 656--672

\bibitem[{Li et~al(2016)Li, Kadav, Durdanovic, Samet, and Graf}]{li}
Li H, Kadav A, Durdanovic I, et~al (2016) Pruning filters for efficient
  convnets. CoRR abs/1608.08710.
  \urlprefix\url{http://arxiv.org/abs/1608.08710},
  {\href{https://arxiv.org/abs/1608.08710}{{https://arxiv.org/abs/arXiv:1608.08710}}}

\bibitem[{Liu et~al(2021)Liu, Haghgoo, Chen, Raghunathan, Koh, Sagawa, Liang,
  and Finn}]{liu2021just}
Liu EZ, Haghgoo B, Chen AS, et~al (2021) Just train twice: Improving group
  robustness without training group information. In: International Conference
  on Machine Learning, PMLR, pp 6781--6792

\bibitem[{Louizos et~al(2017)Louizos, Welling, and
  Kingma}]{louizos2018learning}
Louizos C, Welling M, Kingma DP (2017) Learning sparse neural networks through
  $ l\_0 $ regularization. arXiv preprint arXiv:171201312

\bibitem[{Lundberg and Lee(2017)}]{lundberg2017unified}
Lundberg SM, Lee SI (2017) A unified approach to interpreting model
  predictions. Advances in neural information processing systems 30

\bibitem[{Madry et~al(2017)Madry, Makelov, Schmidt, Tsipras, and Vladu}]{madry}
Madry A, Makelov A, Schmidt L, et~al (2017) Towards deep learning models
  resistant to adversarial attacks. \eprint{1706.06083}

\bibitem[{May et~al(2010)May, Maier, and Dandy}]{may2010data}
May RJ, Maier HR, Dandy GC (2010) Data splitting for artificial neural networks
  using som-based stratified sampling. Neural Networks 23(2):283--294

\bibitem[{Mocanu et~al(2017)Mocanu, Mocanu, Stone, Nguyen, Gibescu, and
  Liotta}]{mocanu}
Mocanu DC, Mocanu E, Stone P, et~al (2017) Evolutionary training of sparse
  artificial neural networks: {A} network science perspective. CoRR
  abs/1707.04780. \urlprefix\url{http://arxiv.org/abs/1707.04780},
  {\href{https://arxiv.org/abs/1707.04780}{{https://arxiv.org/abs/arXiv:1707.04780}}}

\bibitem[{Mostafa and Wang(2019)}]{mostafa}
Mostafa H, Wang X (2019) Parameter efficient training of deep convolutional
  neural networks by dynamic sparse reparameterization. In: International
  Conference on Machine Learning, PMLR, pp 4646--4655

\bibitem[{Narang et~al(2017)Narang, Elsen, Diamos, and
  Sengupta}]{narang2017exploring}
Narang S, Elsen E, Diamos G, et~al (2017) Exploring sparsity in recurrent
  neural networks. arXiv preprint arXiv:170405119

\bibitem[{Prakash et~al(2018)Prakash, Moran, Garber, DiLillo, and
  Storer}]{prakash}
Prakash A, Moran N, Garber S, et~al (2018) Deflecting adversarial attacks with
  pixel deflection. In: Proceedings of the IEEE conference on computer vision
  and pattern recognition, pp 8571--8580

\bibitem[{Raghunathan et~al(2018)Raghunathan, Steinhardt, and
  Liang}]{raghunathan2018semidefinite}
Raghunathan A, Steinhardt J, Liang PS (2018) Semidefinite relaxations for
  certifying robustness to adversarial examples. Advances in neural information
  processing systems 31

\bibitem[{Ross and Doshi-Velez(2018)}]{ross2017improving}
Ross A, Doshi-Velez F (2018) Improving the adversarial robustness and
  interpretability of deep neural networks by regularizing their input
  gradients. In: Proceedings of the {AAAI} Conference on Artificial
  Intelligence

\bibitem[{Sagawa et~al(2019)Sagawa, Koh, Hashimoto, and Liang}]{pangwei}
Sagawa S, Koh PW, Hashimoto TB, et~al (2019) Distributionally robust neural
  networks for group shifts: On the importance of regularization for worst-case
  generalization. CoRR abs/1911.08731.
  \urlprefix\url{http://arxiv.org/abs/1911.08731},
  {\href{https://arxiv.org/abs/1911.08731}{{https://arxiv.org/abs/1911.08731}}}

\bibitem[{Savarese et~al(2020)Savarese, Silva, and Maire}]{savarese2020winning}
Savarese P, Silva H, Maire M (2020) Winning the lottery with continuous
  sparsification. Advances in Neural Information Processing Systems
  33:11,380--11,390

\bibitem[{Singh et~al(2018)Singh, Gehr, Mirman, P{\"u}schel, and
  Vechev}]{singh2018fast}
Singh G, Gehr T, Mirman M, et~al (2018) Fast and effective robustness
  certification. NeurIPS 1(4):6

\bibitem[{Staib and Jegelka(2019)}]{DR1}
Staib M, Jegelka S (2019) Distributionally robust optimization and
  generalization in kernel methods. Advances in Neural Information Processing
  Systems 32

\bibitem[{Szegedy et~al(2014)Szegedy, Zaremba, Sutskever, Bruna, Erhan,
  Goodfellow, and Fergus}]{adversarial_robustness}
Szegedy C, Zaremba W, Sutskever I, et~al (2014) Intriguing properties of neural
  networks. In: International Conference on Learning Representations,
  \urlprefix\url{http://arxiv.org/abs/1312.6199}

\bibitem[{Thompson et~al(2020)Thompson, Greenewald, Lee, and
  Manso}]{DBLP:journals/corr/abs-2007-05558}
Thompson NC, Greenewald KH, Lee K, et~al (2020) The computational limits of
  deep learning. CoRR abs/2007.05558.
  \urlprefix\url{https://arxiv.org/abs/2007.05558},
  {\href{https://arxiv.org/abs/2007.05558}{{https://arxiv.org/abs/2007.05558}}}

\bibitem[{Van~Rossum and Drake(2009)}]{python}
Van~Rossum G, Drake FL (2009) Python 3 Reference Manual. CreateSpace, Scotts
  Valley, CA

\bibitem[{Weng et~al(2018)Weng, Zhang, Chen, Song, Hsieh, Daniel, Boning, and
  Dhillon}]{weng2018towards}
Weng L, Zhang H, Chen H, et~al (2018) Towards fast computation of certified
  robustness for relu networks. In: International Conference on Machine
  Learning, PMLR, pp 5276--5285

\bibitem[{Xiao et~al(2017)Xiao, Rasul, and Vollgraf}]{fashionmnist}
Xiao H, Rasul K, Vollgraf R (2017) Fashion-mnist: a novel image dataset for
  benchmarking machine learning algorithms. \eprint{1708.07747}

\bibitem[{Xie et~al(2017)Xie, Wang, Zhang, Ren, and Yuille}]{xie2017mitigating}
Xie C, Wang J, Zhang Z, et~al (2017) Mitigating adversarial effects through
  randomization. \eprint{1711.01991}

\bibitem[{Xu and Goodacre(2018)}]{xu2018splitting}
Xu Y, Goodacre R (2018) On splitting training and validation set: a comparative
  study of cross-validation, bootstrap and systematic sampling for estimating
  the generalization performance of supervised learning. Journal of analysis
  and testing 2(3):249--262

\bibitem[{Yan et~al(2018)Yan, Guo, and Zhang}]{yan2018deep}
Yan Z, Guo Y, Zhang C (2018) Deep defense: Training dnns with improved
  adversarial robustness. Advances in Neural Information Processing Systems 31

\bibitem[{Zhang et~al(2018)Zhang, Weng, Chen, Hsieh, and
  Daniel}]{zhang2018efficient}
Zhang H, Weng TW, Chen PY, et~al (2018) Efficient neural network robustness
  certification with general activation functions. Advances in neural
  information processing systems 31

\bibitem[{Zhuang et~al(2018)Zhuang, Tan, Zhuang, Liu, Guo, Wu, Huang, and
  Zhu}]{zhuang}
Zhuang Z, Tan M, Zhuang B, et~al (2018) Discrimination-aware channel pruning
  for deep neural networks. Advances in neural information processing systems
  31

\end{thebibliography}

\begin{appendices}


\clearpage

\section{Results Tables}
We present the evaluation results for natural accuracy, adversarial accuracy, stability, and sparsity on the test sets across all data sets and methods discussed in the paper. The natural accuracy results can be found in Table \ref{tab:a1}, adversarial accuracy results in Table \ref{tab:a2}, stability results in Table \ref{tab:a3}, and sparsity results in Table \ref{tab:a4}.

\begin{table}[ht]
\centering
\resizebox{\textwidth}{!}{%
\begin{tabular}{|p{0.15cm}l|p{0.7cm}p{0.3cm}p{0.3cm}llllllll|}
\hline
& \textbf{Name} & $\bm{n}$ & $\bm{p}$ &$\bm{k}$& \textbf{DL} & \textbf{Rob.} & \textbf{St.} & \textbf{Sp.} & \textbf{Rob.} & \textbf{St.} & \textbf{St.} & \textbf{HDL} \\
&  &  &  &  &  & &  & & \textbf{+Sp.} & \textbf{+Sp.} & \textbf{+Rob.} &  \\\hline
\cellcolor{blue!15} &        echocardiogram &     131 &    7 &   3 &    40.00 &   42.22 &   42.96 &   40.00 &          42.22 &          42.22 &          37.78 &  36.30 \\
 \cellcolor{blue!15} &           hill-valley &     605 &  100 &   2 &    47.21 &   47.54 &   53.61 &   47.70 &          47.87 &          52.30 &          51.48 &  51.64 \\
 \cellcolor{blue!15} &        planning-relax &     181 &   12 &   2 &    53.51 &   52.97 &   54.05 &   54.05 &          54.05 &          54.05 &          54.59 &  54.05 \\
 \cellcolor{blue!15} &            poker-hand &   25009 &   10 &  10 &    54.87 &   55.24 &   55.09 &   54.89 &          54.61 &          54.41 &          55.19 &  54.17 \\
 \cellcolor{blue!15} &     hill-valley-noise &     605 &  100 &   2 &    56.23 &   53.44 &   59.67 &   57.87 &          51.31 &          53.11 &          53.61 &  50.82 \\
 \cellcolor{blue!15} &                 yeast &    1483 &    8 &  10 &    58.45 &   59.06 &   59.12 &   59.80 &          60.88 &          59.80 &          59.46 &  60.54 \\
 \cellcolor{blue!15} &     haberman-survival &     305 &    3 &   2 &    62.58 &   61.94 &   61.61 &   63.23 &          61.61 &          61.94 &          60.97 &  62.90 \\
 \cellcolor{blue!15} &  glass-identification &     213 &    9 &   6 &    64.65 &   59.53 &   65.58 &   61.40 &          61.40 &          63.72 &          61.40 &  61.40 \\
 \cellcolor{blue!30} &   brst-cancer-ws-prog &     197 &   32 &   2 &    71.00 &   70.50 &   72.50 &   71.50 &          73.00 &          69.50 &          71.50 &  66.50 \\
 \cellcolor{blue!30} &            hayes-roth &     131 &    4 &   3 &    74.81 &   79.26 &   77.78 &   73.33 &          82.22 &          77.78 &          74.81 &  79.26 \\
 \cellcolor{blue!30} &          spectf-heart &      79 &   44 &   2 &    76.25 &   86.25 &   73.75 &   73.75 &          80.00 &          75.00 &          83.75 &  87.50 \\
 \cellcolor{blue!30} &             hepatitis &     154 &   19 &   2 &    78.71 &   80.00 &   77.42 &   78.71 &          79.35 &          79.35 &          77.42 &  79.35 \\
 \cellcolor{blue!30} &   connectionist-bench &     989 &   10 &  11 &    79.19 &   80.30 &   83.54 &   78.08 &          72.32 &          74.44 &          83.23 &  76.67 \\
 \cellcolor{blue!30} &       libras-movement &     359 &   90 &  15 &    79.44 &   82.50 &   80.00 &   75.83 &          80.56 &          77.22 &          79.44 &  80.00 \\
 \cellcolor{blue!30} &   bld-transf-serv-ctr &     747 &    4 &   2 &    79.47 &   79.07 &   79.07 &   79.33 &          77.20 &          78.93 &          79.20 &  79.60 \\
 \cellcolor{blue!45} &   connect-bench-sonar &     207 &   60 &   2 &    83.33 &   86.67 &   89.52 &   86.19 &          85.71 &          88.10 &          83.81 &  86.67 \\
 \cellcolor{blue!45} &    image-segmentation &     209 &   19 &   7 &    83.81 &   87.14 &   84.29 &   83.33 &          89.52 &          84.76 &          87.14 &  83.81 \\
 \cellcolor{blue!45} &                 ecoli &     335 &    7 &   8 &    84.71 &   84.41 &   85.29 &   85.59 &          85.59 &          85.29 &          84.71 &  85.00 \\
 \cellcolor{blue!45} &   qsar-biodegradation &    1054 &   41 &   2 &    85.12 &   85.69 &   85.21 &   84.55 &          84.74 &          85.21 &          84.93 &  84.36 \\
 \cellcolor{blue!45} &            parkinsons &     194 &   21 &   2 &    86.67 &   85.64 &   87.18 &   88.72 &          86.15 &          86.15 &          86.67 &  85.13 \\
 \cellcolor{blue!45} &       magic-gamma-tel &   19019 &   10 &   2 &    87.11 &   87.20 &   86.98 &   87.17 &          86.66 &          86.50 &          87.28 &  86.80 \\
 \cellcolor{blue!60} &    letter-recognition &   19999 &   16 &  26 &    90.26 &   89.57 &   91.15 &   82.63 &          87.60 &          86.31 &          90.84 &  88.79 \\
 \cellcolor{blue!60} &  statlog-proj-landsat &    4434 &   36 &   6 &    90.39 &   90.67 &   90.64 &   90.44 &          89.97 &          90.53 &          90.46 &  90.03 \\
 \cellcolor{blue!60} &     wall-robot-nav-24 &    5455 &   24 &   4 &    92.36 &   92.23 &   92.11 &   92.49 &          92.89 &          92.88 &          92.47 &  93.08 \\
 \cellcolor{blue!60} &              spambase &    4600 &   57 &   2 &    92.73 &   93.36 &   93.42 &   92.94 &          92.81 &          93.03 &          93.12 &  92.83 \\
 \cellcolor{blue!60} &                 seeds &     209 &    7 &   3 &    92.86 &   92.38 &   91.43 &   93.33 &          93.33 &          92.86 &          92.38 &  93.33 \\
 \cellcolor{blue!60} &     ozone-level-eight &    2533 &   72 &   2 &    93.69 &   93.06 &   93.89 &   93.37 &          93.73 &          93.49 &          93.69 &  93.96 \\
 \cellcolor{blue!60} &                cnae-9 &    1079 &  856 &   9 &    93.70 &   93.98 &   94.07 &   92.87 &          93.24 &          92.59 &          93.89 &  92.87 \\
 \cellcolor{blue!60} &         balance-scale &     624 &    4 &   3 &    94.24 &   92.96 &   93.60 &   91.68 &          88.64 &          93.12 &          94.24 &  94.56 \\
 \cellcolor{blue!60} &            ionosphere &     350 &   34 &   2 &    94.93 &   94.93 &   94.37 &   95.21 &          91.83 &          93.24 &          95.77 &  94.08 \\
 \cellcolor{blue!75} &   brst-cancer-ws-orig &     698 &    9 &   2 &    96.00 &   96.29 &   95.86 &   96.14 &          96.86 &          96.00 &          96.43 &  96.29 \\
 \cellcolor{blue!75} &   brst-cancer-ws-diag &     568 &   30 &   2 &    97.37 &   96.49 &   97.37 &   96.49 &          96.49 &          96.14 &          97.19 &  96.67 \\
 \cellcolor{blue!75} &       ozone-level-one &    2535 &   72 &   2 &    97.40 &   97.13 &   97.20 &   97.01 &          97.44 &          97.44 &          97.28 &  97.28 \\
 \cellcolor{blue!75} &      wall-robot-nav-4 &    5455 &    4 &   4 &    97.77 &   97.82 &   97.69 &   98.04 &          98.44 &          98.13 &          97.80 &  98.33 \\
 \cellcolor{blue!75} &    climate-simu-crash &     539 &   18 &   2 &    97.78 &   97.59 &   97.78 &   96.67 &          96.85 &          96.67 &          97.59 &  97.04 \\
 \cellcolor{blue!75} &  optical-recog-digits &    3822 &   64 &  10 &    97.83 &   98.01 &   97.73 &   97.59 &          98.09 &          98.07 &          98.14 &  97.93 \\
 \cellcolor{blue!75} &      wall-robot-nav-2 &    5455 &    2 &   4 &    97.88 &   98.13 &   98.02 &   98.30 &          98.13 &          98.30 &          98.33 &  98.39 \\
 \cellcolor{blue!95} &           dermatology &     365 &   34 &   6 &    98.38 &   98.38 &   98.38 &   98.65 &          98.38 &          98.38 &          98.38 &  98.92 \\
 \cellcolor{blue!95} &   thyroid-disease-new &     214 &    5 &   3 &    98.60 &   99.07 &   98.14 &   98.60 &          99.07 &          96.74 &          99.07 &  97.67 \\
 \cellcolor{blue!95} &   thyroid-disease-ann &    3771 &   21 &   3 &    98.86 &   98.91 &   98.78 &   98.70 &          98.89 &          98.86 &          99.05 &  98.91 \\
 \cellcolor{blue!95} &                  wine &     177 &   13 &   3 &    98.89 &   98.33 &   98.89 &  100.00 &          97.78 &          97.22 &          97.78 &  98.33 \\
 \cellcolor{blue!95} &      pen-recog-digits &    7493 &   16 &  10 &    99.23 &   99.05 &   99.16 &   99.11 &          99.41 &          99.31 &          99.19 &  99.41 \\
 \cellcolor{blue!95} &     skin-segmentation &  245056 &    3 &   2 &    99.91 &   99.90 &   99.90 &   99.90 &          99.88 &          99.89 &          99.91 &  99.89 \\
 \cellcolor{blue!95} &      banknote-authent &    1371 &    4 &   2 &    99.93 &  100.00 &  100.00 &   99.85 &          97.16 &          89.75 &         100.00 &  95.71 \\
 \cellcolor{blue!95} &                  iris &     149 &    4 &   3 &   100.00 &   99.33 &   98.00 &  100.00 &          92.67 &          98.67 &          99.33 &  99.33 \\
\hline
\cellcolor{red!30}&MNIST &    70000 &   784 &  10 &    99.19 &     99.29 &   99.21 &          99.13 &          99.22 &   99.16    &   99.31 &  99.30 \\
\cellcolor{red!30}&Fashion-MNIST &    70000 &   784 &  10 &    91.77 &     91.88 &   91.93 &          91.54 &          92.12 &   91.43    &   92.00 &  92.15 \\
\cellcolor{red!30}&CIFAR10 &    60000 &   1024 &  10 &    68.42 &     69.15 &   68.25 &          66.60 &          57.74 &   69.82    &   68.78 &  69.03 \\
\hline
\end{tabular}%
}
\caption{Natural accuracy results for all UCI and vision data sets, where $n$ denotes the data size, $p$ denotes the data dimension, and $k$ denotes the number of classes. Darker blue corresponds to higher nominal DL natural accuracy for the UCI data sets.}\label{tab:a1}
\end{table}

\begin{table}[t]
\centering
\resizebox{\textwidth}{!}{%
\begin{tabular}{|p{0.15cm}l|p{0.7cm}p{0.3cm}p{0.3cm}llllllll|}
\hline
& \textbf{Name} & $\bm{n}$ & $\bm{p}$ &$\bm{k}$& \textbf{DL} & \textbf{Rob.} & \textbf{St.} & \textbf{Sp.} & \textbf{Rob.} & \textbf{St.} & \textbf{St.} & \textbf{HDL} \\
&  &  &  &  &  & &  & & \textbf{+Sp.} & \textbf{+Sp.} & \textbf{+Rob.} &  \\\hline
 \cellcolor{blue!15} &        echocardiogram &     131 &    7 &   3 &     4.44 &   44.44 &    7.41 &   17.78 &          44.44 &          17.78 &          41.48 &  44.44 \\
 \cellcolor{blue!15} &           hill-valley &     605 &  100 &   2 &     7.54 &   40.16 &   28.52 &   20.00 &          39.02 &          20.16 &          38.52 &  36.39 \\
 \cellcolor{blue!15} &        planning-relax &     181 &   12 &   2 &    21.62 &   54.05 &   44.86 &   49.19 &          54.05 &          54.05 &          54.05 &  54.05 \\
 \cellcolor{blue!15} &            poker-hand &   25009 &   10 &  10 &    50.70 &   50.70 &   50.70 &   50.70 &          50.70 &          50.70 &          50.70 &  51.95 \\
 \cellcolor{blue!15} &     hill-valley-noise &     605 &  100 &   2 &    11.31 &   21.48 &   22.30 &   12.79 &          30.00 &          20.98 &          24.10 &  34.43 \\
 \cellcolor{blue!15} &                 yeast &    1483 &    8 &  10 &     0.47 &   32.26 &    0.27 &    0.20 &          32.26 &           0.00 &          32.26 &  32.26 \\
 \cellcolor{blue!15} &     haberman-survival &     305 &    3 &   2 &    17.74 &   59.68 &   20.97 &   47.74 &          59.68 &          34.19 &          59.68 &  59.68 \\
 \cellcolor{blue!15} &  glass-identification &     213 &    9 &   6 &     4.65 &   20.93 &    5.58 &    5.12 &          20.93 &           9.30 &          25.58 &  25.58 \\
 \cellcolor{blue!30} &   brst-cancer-ws-prog &     197 &   32 &   2 &    18.00 &   72.50 &   21.00 &   43.50 &          72.50 &          39.50 &          72.50 &  72.50 \\
 \cellcolor{blue!30} &            hayes-roth &     131 &    4 &   3 &     9.63 &   32.59 &    8.89 &    5.19 &          36.30 &          13.33 &          40.74 &  40.74 \\
 \cellcolor{blue!30} &          spectf-heart &      79 &   44 &   2 &    27.50 &   25.00 &   32.50 &   40.00 &          25.00 &          13.75 &          21.25 &  25.00 \\
 \cellcolor{blue!30} &             hepatitis &     154 &   19 &   2 &     5.81 &   70.97 &    9.03 &   21.94 &          70.97 &          29.68 &          70.97 &  70.97 \\
 \cellcolor{blue!30} &   connectionist-bench &     989 &   10 &  11 &     4.04 &    1.52 &    4.34 &    1.41 &           6.46 &           3.84 &           8.79 &   8.99 \\
 \cellcolor{blue!30} &       libras-movement &     359 &   90 &  15 &     0.00 &    0.83 &    0.28 &    0.00 &           2.50 &           0.00 &           4.72 &   2.50 \\
 \cellcolor{blue!30} &   bld-transf-serv-ctr &     747 &    4 &   2 &     0.00 &   72.67 &    0.13 &   17.60 &          72.67 &          43.60 &          72.67 &  72.67 \\
 \cellcolor{blue!45} &   connect-bench-sonar &     207 &   60 &   2 &     5.71 &   20.00 &   18.10 &   32.38 &          43.81 &          33.81 &          22.86 &  40.00 \\
 \cellcolor{blue!45} &    image-segmentation &     209 &   19 &   7 &     4.29 &    0.95 &    2.86 &    2.38 &           9.52 &           3.33 &           9.52 &   8.10 \\
 \cellcolor{blue!45} &                 ecoli &     335 &    7 &   8 &     0.59 &   41.18 &    1.76 &    2.06 &          51.47 &           0.88 &          41.18 &  51.47 \\
 \cellcolor{blue!45} &   qsar-biodegradation &    1054 &   41 &   2 &    36.21 &   72.04 &   24.08 &   31.75 &          72.04 &          29.67 &          72.04 &  72.04 \\
 \cellcolor{blue!45} &            parkinsons &     194 &   21 &   2 &    58.97 &   74.36 &   63.08 &   63.08 &          74.36 &          68.72 &          74.36 &  74.36 \\
 \cellcolor{blue!45} &       magic-gamma-tel &   19019 &   10 &   2 &    15.07 &   64.59 &   15.28 &   18.62 &          64.59 &          10.36 &          64.59 &  64.59 \\
 \cellcolor{blue!60} &    letter-recognition &   19999 &   16 &  26 &     0.76 &    3.73 &    1.11 &    0.52 &           3.68 &           0.34 &           3.57 &   3.74 \\
 \cellcolor{blue!60} &  statlog-proj-landsat &    4434 &   36 &   6 &     8.66 &   25.48 &   10.35 &    7.51 &          25.48 &           5.77 &          20.79 &  25.39 \\
 \cellcolor{blue!60} &     wall-robot-nav-24 &    5455 &   24 &   4 &     3.04 &   39.69 &    3.28 &    1.63 &          39.69 &           3.06 &          39.69 &  39.65 \\
 \cellcolor{blue!60} &              spambase &    4600 &   57 &   2 &    40.67 &   58.41 &   48.08 &   45.06 &          58.41 &          49.25 &          58.41 &  58.41 \\
 \cellcolor{blue!60} &                 seeds &     209 &    7 &   3 &    23.81 &   24.29 &   31.90 &   27.14 &          32.38 &          15.24 &          31.43 &  30.95 \\
 \cellcolor{blue!60} &     ozone-level-eight &    2533 &   72 &   2 &    49.59 &   94.48 &   48.88 &   94.48 &          94.48 &          56.69 &          94.48 &  94.48 \\
 \cellcolor{blue!60} &                cnae-9 &    1079 &  856 &   9 &     0.00 &    1.02 &    0.09 &    0.83 &           1.94 &           2.50 &           3.80 &   3.80 \\
 \cellcolor{blue!60} &         balance-scale &     624 &    4 &   3 &    17.44 &   40.80 &   13.28 &    8.16 &          43.84 &          14.72 &          49.92 &  49.92 \\
 \cellcolor{blue!60} &            ionosphere &     350 &   34 &   2 &    19.44 &   57.75 &   25.35 &   34.65 &          57.75 &          10.99 &          57.75 &  57.75 \\
 \cellcolor{blue!75} &   brst-cancer-ws-orig &     698 &    9 &   2 &    17.71 &   60.71 &   23.71 &   31.29 &          60.71 &          15.14 &          60.71 &  60.71 \\
 \cellcolor{blue!75} &   brst-cancer-ws-diag &     568 &   30 &   2 &    25.44 &   47.02 &   25.61 &   47.02 &          58.77 &          13.33 &          47.02 &  58.77 \\
 \cellcolor{blue!75} &       ozone-level-one &    2535 &   72 &   2 &    73.78 &   97.44 &   81.22 &   97.44 &          97.44 &          89.17 &          97.44 &  97.44 \\
 \cellcolor{blue!75} &      wall-robot-nav-4 &    5455 &    4 &   4 &     3.42 &   39.69 &    4.85 &    7.69 &          39.69 &          10.49 &          39.69 &  39.69 \\
 \cellcolor{blue!75} &    climate-simu-crash &     539 &   18 &   2 &     0.00 &   94.44 &    2.41 &   75.56 &          94.44 &          83.33 &          94.44 &  94.44 \\
 \cellcolor{blue!75} &  optical-recog-digits &    3822 &   64 &  10 &     1.36 &    7.48 &    1.39 &    0.76 &           7.16 &           0.60 &           8.94 &  10.27 \\
 \cellcolor{blue!75} &      wall-robot-nav-2 &    5455 &    2 &   4 &     5.26 &   39.69 &    5.48 &   17.75 &          39.69 &          21.03 &          39.69 &  39.69 \\
 \cellcolor{blue!95} &           dermatology &     365 &   34 &   6 &     4.59 &   20.27 &    2.97 &    5.68 &          20.27 &          10.81 &          20.27 &  20.27 \\
 \cellcolor{blue!95} &   thyroid-disease-new &     214 &    5 &   3 &     9.77 &   65.12 &   10.23 &   13.02 &          65.12 &          13.02 &          65.12 &  65.12 \\
 \cellcolor{blue!95} &   thyroid-disease-ann &    3771 &   21 &   3 &    48.42 &   91.79 &   54.97 &   54.12 &          91.79 &          55.23 &          91.79 &  91.79 \\
 \cellcolor{blue!95} &                  wine &     177 &   13 &   3 &     1.67 &   44.44 &    1.11 &    3.89 &          37.78 &           5.56 &          43.33 &  31.67 \\
 \cellcolor{blue!95} &      pen-recog-digits &    7493 &   16 &  10 &     2.76 &   10.13 &    2.63 &    2.05 &          10.13 &           1.80 &           8.71 &  10.27 \\
 \cellcolor{blue!95} &     skin-segmentation &  245056 &    3 &   2 &    79.30 &   79.30 &   79.30 &   79.30 &          79.30 &          79.28 &          79.30 &  79.30 \\
 \cellcolor{blue!95} &      banknote-authent &    1371 &    4 &   2 &    39.20 &   54.25 &   42.25 &   54.25 &          54.25 &          43.13 &          54.25 &  54.25 \\
 \cellcolor{blue!95} &                  iris &     149 &    4 &   3 &    12.00 &   16.00 &   12.00 &   10.00 &          24.67 &          10.00 &          32.67 &  32.67 \\
\hline
\cellcolor{red!30}&MNIST &    70000 &   784 &  10 &    49.60 &     78.37 &   51.52 &          43.80 &          74.67 &   39.94    &   79.50 &  75.97 \\
\cellcolor{red!30}&Fashion-MNIST &    70000 &   784 &  10 &    78.70 &     87.74 &   78.30 &          80.76 &          87.07 &   80.24    &   87.80 &  86.91 \\
\cellcolor{red!30}&CIFAR10 &    60000 &   1024 &  10 &       28.81 &   48.61 &      28.75 &          34.87 &   43.98    &   33.73 & 47.32 & 43.96 \\
\hline
\end{tabular}%
}
\caption{Adversarial accuracy results for all UCI and vision data sets, where $n$ denotes the data size, $p$ denotes the data dimension, and $k$ denotes the number of classes. We use $\rho =0.1$ for all data sets except CIFAR10 and Fashion-MNIST, for which we set $\rho =0.01$. Darker blue corresponds to higher nominal (DL) natural accuracy.}\label{tab:a2}
\end{table}

\begin{table}[t]
\centering
\resizebox{\textwidth}{!}{%
\begin{tabular}{|p{0.15cm}l|p{0.7cm}p{0.3cm}p{0.3cm}llllllll|}
\hline
& \textbf{Name} & $\bm{n}$ & $\bm{p}$ &$\bm{k}$& \textbf{DL} & \textbf{Rob.} & \textbf{St.} & \textbf{Sp.} & \textbf{Rob.} & \textbf{St.} & \textbf{St.} & \textbf{HDL} \\
&  &  &  &  &  & &  & & \textbf{+Sp.} & \textbf{+Sp.} & \textbf{+Rob.} &  \\\hline
 \cellcolor{blue!15} &        echocardiogram &     131 &    7 &   3 &    40.74 &   37.04 &   37.04 &   40.74 &          33.33 &          33.33 &          40.74 &  33.33 \\
 \cellcolor{blue!15} &           hill-valley &     605 &  100 &   2 &    43.44 &   43.44 &   49.18 &   43.44 &          46.72 &          50.82 &          45.08 &  49.18 \\
 \cellcolor{blue!15} &        planning-relax &     181 &   12 &   2 &    51.35 &   48.65 &   54.05 &   54.05 &          54.05 &          51.35 &          54.05 &  54.05 \\
 \cellcolor{blue!15} &            poker-hand &   25009 &   10 &  10 &    54.60 &   54.32 &   54.56 &   54.48 &          53.20 &          53.52 &          54.70 &  52.32 \\
 \cellcolor{blue!15} &     hill-valley-noise &     605 &  100 &   2 &    54.92 &   47.54 &   52.46 &   54.92 &          47.54 &          47.54 &          48.36 &  50.00 \\
 \cellcolor{blue!15} &                 yeast &    1483 &    8 &  10 &    57.58 &   56.90 &   56.57 &   57.24 &          59.60 &          58.92 &          56.90 &  58.92 \\
 \cellcolor{blue!15} &     haberman-survival &     305 &    3 &   2 &    59.68 &   59.68 &   59.68 &   59.68 &          59.68 &          59.68 &          59.68 &  59.68 \\
 \cellcolor{blue!15} &  glass-identification &     213 &    9 &   6 &    55.81 &   58.14 &   53.49 &   58.14 &          55.81 &          58.14 &          58.14 &  58.14 \\
 \cellcolor{blue!30} &   brst-cancer-ws-prog &     197 &   32 &   2 &    67.50 &   67.50 &   67.50 &   67.50 &          62.50 &          60.00 &          72.50 &  67.50 \\
 \cellcolor{blue!30} &            hayes-roth &     131 &    4 &   3 &    70.37 &   74.07 &   70.37 &   70.37 &          70.37 &          74.07 &          70.37 &  74.07 \\
 \cellcolor{blue!30} &          spectf-heart &      79 &   44 &   2 &    68.75 &   68.75 &   75.00 &   68.75 &          68.75 &          62.50 &          75.00 &  75.00 \\
 \cellcolor{blue!30} &             hepatitis &     154 &   19 &   2 &    70.97 &   70.97 &   70.97 &   70.97 &          74.19 &          70.97 &          70.97 &  77.42 \\
 \cellcolor{blue!30} &   connectionist-bench &     989 &   10 &  11 &    75.25 &   77.78 &   80.81 &   71.72 &          63.13 &          60.61 &          76.26 &  70.71 \\
 \cellcolor{blue!30} &       libras-movement &     359 &   90 &  15 &    75.00 &   75.00 &   79.17 &   68.06 &          75.00 &          70.83 &          81.94 &  75.00 \\
 \cellcolor{blue!30} &   bld-transf-serv-ctr &     747 &    4 &   2 &    79.33 &   77.33 &   78.00 &   78.67 &          74.67 &          74.00 &          78.00 &  79.33 \\
 \cellcolor{blue!45} &   connect-bench-sonar &     207 &   60 &   2 &    78.57 &   80.95 &   83.33 &   80.95 &          83.33 &          83.33 &          80.95 &  83.33 \\
 \cellcolor{blue!45} &    image-segmentation &     209 &   19 &   7 &    78.57 &   78.57 &   80.95 &   76.19 &          73.81 &          83.33 &          78.57 &  78.57 \\
 \cellcolor{blue!45} &                 ecoli &     335 &    7 &   8 &    80.88 &   77.94 &   83.82 &   82.35 &          83.82 &          83.82 &          83.82 &  82.35 \\
 \cellcolor{blue!45} &   qsar-biodegradation &    1054 &   41 &   2 &    84.83 &   84.83 &   83.89 &   81.99 &          83.41 &          83.89 &          84.36 &  83.89 \\
 \cellcolor{blue!45} &            parkinsons &     194 &   21 &   2 &    84.62 &   82.05 &   82.05 &   84.62 &          79.49 &          76.92 &          84.62 &  79.49 \\
 \cellcolor{blue!45} &       magic-gamma-tel &   19019 &   10 &   2 &    86.65 &   86.93 &   86.44 &   86.75 &          85.73 &          86.01 &          87.01 &  86.25 \\
 \cellcolor{blue!60} &    letter-recognition &   19999 &   16 &  26 &    86.98 &   86.75 &   89.60 &   82.27 &          84.47 &          83.85 &          90.20 &  86.55 \\
 \cellcolor{blue!60} &  statlog-proj-landsat &    4434 &   36 &   6 &    88.84 &   88.05 &   90.08 &   90.19 &          88.61 &          89.40 &          89.29 &  88.39 \\
 \cellcolor{blue!60} &     wall-robot-nav-24 &    5455 &   24 &   4 &    92.03 &   91.12 &   91.58 &   92.31 &          92.03 &          92.22 &          90.75 &  92.40 \\
 \cellcolor{blue!60} &              spambase &    4600 &   57 &   2 &    92.18 &   92.73 &   92.62 &   92.29 &          92.40 &          92.51 &          92.40 &  92.62 \\
 \cellcolor{blue!60} &                 seeds &     209 &    7 &   3 &    90.48 &   90.48 &   92.86 &   92.86 &          92.86 &          90.48 &          90.48 &  92.86 \\
 \cellcolor{blue!60} &     ozone-level-eight &    2533 &   72 &   2 &    93.10 &   94.08 &   93.10 &   93.29 &          94.48 &          92.31 &          93.10 &  93.49 \\
 \cellcolor{blue!60} &                cnae-9 &    1079 &  856 &   9 &    91.20 &   93.06 &   92.59 &   92.13 &          92.59 &          92.13 &          93.06 &  91.20 \\
 \cellcolor{blue!60} &         balance-scale &     624 &    4 &   3 &    91.20 &   91.20 &   92.00 &   90.40 &          72.80 &          91.20 &          91.20 &  92.80 \\
 \cellcolor{blue!60} &            ionosphere &     350 &   34 &   2 &    92.96 &   90.14 &   92.96 &   91.55 &          94.37 &          91.55 &          90.14 &  91.55 \\
 \cellcolor{blue!75} &   brst-cancer-ws-orig &     698 &    9 &   2 &    93.57 &   93.57 &   95.71 &   95.71 &          95.71 &          95.00 &          95.71 &  95.71 \\
 \cellcolor{blue!75} &   brst-cancer-ws-diag &     568 &   30 &   2 &    95.61 &   94.74 &   95.61 &   94.74 &          95.61 &          93.86 &          94.74 &  95.61 \\
 \cellcolor{blue!75} &       ozone-level-one &    2535 &   72 &   2 &    97.24 &   96.46 &   96.85 &   96.85 &          97.44 &          97.44 &          96.85 &  97.05 \\
 \cellcolor{blue!75} &      wall-robot-nav-4 &    5455 &    4 &   4 &    97.44 &   97.16 &   97.34 &   97.25 &          97.71 &          97.62 &          97.62 &  97.80 \\
 \cellcolor{blue!75} &    climate-simu-crash &     539 &   18 &   2 &    96.30 &   96.30 &   95.37 &   94.44 &          96.30 &          93.52 &          96.30 &  96.30 \\
 \cellcolor{blue!75} &  optical-recog-digits &    3822 &   64 &  10 &    97.39 &   97.39 &   96.99 &   97.12 &          97.91 &          97.91 &          97.78 &  97.65 \\
 \cellcolor{blue!75} &      wall-robot-nav-2 &    5455 &    2 &   4 &    96.98 &   97.53 &   97.34 &   97.25 &          97.53 &          98.17 &          97.89 &  97.44 \\
 \cellcolor{blue!95} &           dermatology &     365 &   34 &   6 &    97.30 &   97.30 &   97.30 &   98.65 &          95.95 &          97.30 &          97.30 &  94.59 \\
 \cellcolor{blue!95} &   thyroid-disease-new &     214 &    5 &   3 &    97.67 &   97.67 &   95.35 &   93.02 &          95.35 &          93.02 &          97.67 &  93.02 \\
 \cellcolor{blue!95} &   thyroid-disease-ann &    3771 &   21 &   3 &    98.28 &   98.81 &   98.01 &   98.54 &          98.68 &          98.41 &          98.54 &  98.28 \\
 \cellcolor{blue!95} &                  wine &     177 &   13 &   3 &    94.44 &   97.22 &   94.44 &   94.44 &          97.22 &          88.89 &          97.22 &  97.22 \\
 \cellcolor{blue!95} &      pen-recog-digits &    7493 &   16 &  10 &    98.93 &   98.67 &   99.00 &   99.07 &          99.13 &          99.33 &          99.07 &  99.07 \\
 \cellcolor{blue!95} &     skin-segmentation &  245056 &    3 &   2 &    99.90 &   99.89 &   99.89 &   99.89 &          99.86 &          99.87 &          99.90 &  99.87 \\
 \cellcolor{blue!95} &      banknote-authent &    1371 &    4 &   2 &    99.64 &  100.00 &  100.00 &   99.64 &          87.27 &          70.18 &         100.00 &  87.27 \\
 \cellcolor{blue!95} &                  iris &     149 &    4 &   3 &   100.00 &  100.00 &   93.33 &  100.00 &          70.00 &          96.67 &          96.67 &  96.67 \\
\hline
\cellcolor{red!30}&MNIST &    70000 &   784 &  10 &    99.14 &     99.24 &   99.15 &          99.06 &          99.07 &   98.86    &   99.25 &  99.19 \\
\cellcolor{red!30}&Fashion-MNIST &    70000 &   784 &  10 &    91.53 &     91.36 &   91.75 &          91.29&          91.65 &   91.15    &  91.38 &  90.19 \\
\cellcolor{red!30}&CIFAR10 &    60000 &   1024 &  10 &       68.28 &   65.44 &      68.13 &          63.14 &   56.04    &   61.76 & 63.76 & 56.48 \\
\hline
\end{tabular}%
}
\caption{Stability (worst case accuracy) results for all UCI and vision data sets, where $n$ denotes the data size, $p$ denotes the data dimension, and $k$ denotes the number of classes. Darker blue corresponds to higher nominal (DL) natural accuracy.}\label{tab:a3}
\end{table}

\begin{table}[t]
\centering
\resizebox{\textwidth}{!}{%
\begin{tabular}{|p{0.15cm}l|p{0.7cm}p{0.3cm}p{0.3cm}llllllll|}
\hline
& \textbf{Name} & $\bm{n}$ & $\bm{p}$ &$\bm{k}$& \textbf{DL} & \textbf{Rob.} & \textbf{St.} & \textbf{Sp.} & \textbf{Rob.} & \textbf{St.} & \textbf{St.} & \textbf{HDL} \\
&  &  &  &  &  & &  & & \textbf{+Sp.} & \textbf{+Sp.} & \textbf{+Rob.} &  \\\hline
\cellcolor{blue!15} &        echocardiogram &     131 &    7 &   3 &      0.0 &     0.0 &     0.0 &   52.17 &          62.57 &          65.69 &            0.0 &  65.49 \\
 \cellcolor{blue!15} &           hill-valley &     605 &  100 &   2 &      0.0 &     0.0 &     0.0 &   49.00 &          41.94 &          34.89 &            0.0 &  28.80 \\
 \cellcolor{blue!15} &        planning-relax &     181 &   12 &   2 &      0.0 &     0.0 &     0.0 &   54.47 &          63.00 &          69.13 &            0.0 &  70.56 \\
 \cellcolor{blue!15} &            poker-hand &   25009 &   10 &  10 &      0.0 &     0.0 &     0.0 &   29.56 &          49.23 &          51.31 &            0.0 &  49.83 \\
 \cellcolor{blue!15} &     hill-valley-noise &     605 &  100 &   2 &      0.0 &     0.0 &     0.0 &   50.67 &          39.73 &          45.04 &            0.0 &  29.28 \\
 \cellcolor{blue!15} &                 yeast &    1483 &    8 &  10 &      0.0 &     0.0 &     0.0 &   53.08 &          53.41 &          54.71 &            0.0 &  54.36 \\
 \cellcolor{blue!15} &     haberman-survival &     305 &    3 &   2 &      0.0 &     0.0 &     0.0 &   47.21 &          49.40 &          53.34 &            0.0 &  53.10 \\
 \cellcolor{blue!15} &  glass-identification &     213 &    9 &   6 &      0.0 &     0.0 &     0.0 &   56.14 &          63.65 &          64.86 &            0.0 &  64.85 \\
 \cellcolor{blue!30} &   brst-cancer-ws-prog &     197 &   32 &   2 &      0.0 &     0.0 &     0.0 &   61.87 &          67.34 &          69.49 &            0.0 &  67.78 \\
 \cellcolor{blue!30} &            hayes-roth &     131 &    4 &   3 &      0.0 &     0.0 &     0.0 &   61.56 &          67.49 &          66.39 &            0.0 &  67.38 \\
 \cellcolor{blue!30} &          spectf-heart &      79 &   44 &   2 &      0.0 &     0.0 &     0.0 &   81.02 &          85.84 &          86.10 &            0.0 &  85.58 \\
 \cellcolor{blue!30} &             hepatitis &     154 &   19 &   2 &      0.0 &     0.0 &     0.0 &   64.56 &          68.71 &          74.01 &            0.0 &  72.74 \\
 \cellcolor{blue!30} &   connectionist-bench &     989 &   10 &  11 &      0.0 &     0.0 &     0.0 &   57.95 &          63.20 &          63.34 &            0.0 &  63.45 \\
 \cellcolor{blue!30} &       libras-movement &     359 &   90 &  15 &      0.0 &     0.0 &     0.0 &   66.76 &          70.90 &          71.00 &            0.0 &  70.73 \\
 \cellcolor{blue!30} &   bld-transf-serv-ctr &     747 &    4 &   2 &      0.0 &     0.0 &     0.0 &   46.21 &          48.90 &          52.65 &            0.0 &  50.52 \\
 \cellcolor{blue!45} &   connect-bench-sonar &     207 &   60 &   2 &      0.0 &     0.0 &     0.0 &   76.09 &          80.04 &          81.10 &            0.0 &  80.07 \\
 \cellcolor{blue!45} &    image-segmentation &     209 &   19 &   7 &      0.0 &     0.0 &     0.0 &   60.95 &          66.47 &          69.23 &            0.0 &  67.24 \\
 \cellcolor{blue!45} &                 ecoli &     335 &    7 &   8 &      0.0 &     0.0 &     0.0 &   54.91 &          62.25 &          63.26 &            0.0 &  62.31 \\
 \cellcolor{blue!45} &   qsar-biodegradation &    1054 &   41 &   2 &      0.0 &     0.0 &     0.0 &   44.89 &          44.32 &          51.82 &            0.0 &  47.65 \\
 \cellcolor{blue!45} &            parkinsons &     194 &   21 &   2 &      0.0 &     0.0 &     0.0 &   59.42 &          60.95 &          67.40 &            0.0 &  63.23 \\
 \cellcolor{blue!45} &       magic-gamma-tel &   19019 &   10 &   2 &      0.0 &     0.0 &     0.0 &   50.80 &          51.43 &          57.52 &            0.0 &  52.84 \\
 \cellcolor{blue!60} &    letter-recognition &   19999 &   16 &  26 &      0.0 &     0.0 &     0.0 &   58.77 &          63.84 &          65.32 &            0.0 &  64.86 \\
 \cellcolor{blue!60} &  statlog-proj-landsat &    4434 &   36 &   6 &      0.0 &     0.0 &     0.0 &   45.65 &          50.60 &          52.06 &            0.0 &  51.53 \\
 \cellcolor{blue!60} &     wall-robot-nav-24 &    5455 &   24 &   4 &      0.0 &     0.0 &     0.0 &   51.85 &          55.96 &          56.59 &            0.0 &  55.53 \\
 \cellcolor{blue!60} &              spambase &    4600 &   57 &   2 &      0.0 &     0.0 &     0.0 &   56.81 &          56.56 &          59.90 &            0.0 &  55.84 \\
 \cellcolor{blue!60} &                 seeds &     209 &    7 &   3 &      0.0 &     0.0 &     0.0 &   65.56 &          71.45 &          72.73 &            0.0 &  72.33 \\
 \cellcolor{blue!60} &     ozone-level-eight &    2533 &   72 &   2 &      0.0 &     0.0 &     0.0 &   66.09 &          68.13 &          67.86 &            0.0 &  67.38 \\
 \cellcolor{blue!60} &                cnae-9 &    1079 &  856 &   9 &      0.0 &     0.0 &     0.0 &   61.94 &          62.06 &          73.43 &            0.0 &  61.91 \\
 \cellcolor{blue!60} &         balance-scale &     624 &    4 &   3 &      0.0 &     0.0 &     0.0 &   57.25 &          63.41 &          63.35 &            0.0 &  63.14 \\
 \cellcolor{blue!60} &            ionosphere &     350 &   34 &   2 &      0.0 &     0.0 &     0.0 &   63.16 &          66.76 &          68.17 &            0.0 &  66.74 \\
 \cellcolor{blue!75} &   brst-cancer-ws-orig &     698 &    9 &   2 &      0.0 &     0.0 &     0.0 &   48.29 &          53.44 &          55.78 &            0.0 &  55.97 \\
 \cellcolor{blue!75} &   brst-cancer-ws-diag &     568 &   30 &   2 &      0.0 &     0.0 &     0.0 &   68.91 &          71.38 &          71.46 &            0.0 &  70.69 \\
 \cellcolor{blue!75} &       ozone-level-one &    2535 &   72 &   2 &      0.0 &     0.0 &     0.0 &   66.22 &          68.25 &          67.97 &            0.0 &  68.58 \\
 \cellcolor{blue!75} &      wall-robot-nav-4 &    5455 &    4 &   4 &      0.0 &     0.0 &     0.0 &   52.89 &          61.43 &          60.75 &            0.0 &  60.50 \\
 \cellcolor{blue!75} &    climate-simu-crash &     539 &   18 &   2 &      0.0 &     0.0 &     0.0 &   59.75 &          64.56 &          65.88 &            0.0 &  66.29 \\
 \cellcolor{blue!75} &  optical-recog-digits &    3822 &   64 &  10 &      0.0 &     0.0 &     0.0 &   65.42 &          68.85 &          71.25 &            0.0 &  69.22 \\
 \cellcolor{blue!75} &      wall-robot-nav-2 &    5455 &    2 &   4 &      0.0 &     0.0 &     0.0 &   63.85 &          72.12 &          72.68 &            0.0 &  71.74 \\
 \cellcolor{blue!95} &           dermatology &     365 &   34 &   6 &      0.0 &     0.0 &     0.0 &   67.03 &          73.49 &          74.61 &            0.0 &  74.21 \\
 \cellcolor{blue!95} &   thyroid-disease-new &     214 &    5 &   3 &      0.0 &     0.0 &     0.0 &   67.25 &          74.48 &          74.62 &            0.0 &  74.12 \\
 \cellcolor{blue!95} &   thyroid-disease-ann &    3771 &   21 &   3 &      0.0 &     0.0 &     0.0 &   48.81 &          50.15 &          53.22 &            0.0 &  50.36 \\
 \cellcolor{blue!95} &                  wine &     177 &   13 &   3 &      0.0 &     0.0 &     0.0 &   74.62 &          81.05 &          80.60 &            0.0 &  80.59 \\
 \cellcolor{blue!95} &      pen-recog-digits &    7493 &   16 &  10 &      0.0 &     0.0 &     0.0 &   59.13 &          62.58 &          63.33 &            0.0 &  63.23 \\
 \cellcolor{blue!95} &     skin-segmentation &  245056 &    3 &   2 &      0.0 &     0.0 &     0.0 &   63.77 &          71.27 &          72.15 &            0.0 &  72.56 \\
 \cellcolor{blue!95} &      banknote-authent &    1371 &    4 &   2 &      0.0 &     0.0 &     0.0 &   76.88 &          84.17 &          84.97 &            0.0 &  84.98 \\
 \cellcolor{blue!95} &                  iris &     149 &    4 &   3 &      0.0 &     0.0 &     0.0 &   69.89 &          78.03 &          77.56 &            0.0 &  77.79 \\
\hline
 \cellcolor{red!30} &          MNIST &  70000 &   784 &  10 &      0.0 &     0.0 &     0.0 &   39.20 &          75.06 &          44.77 &            0.0 &  76.05 \\
 \cellcolor{red!30} &  Fashion-MNIST &  70000 &   784 &  10 &      0.0 &     0.0 &     0.0 &   45.41 &          68.94 &          48.40 &            0.0 &  69.18 \\
 \cellcolor{red!30} &          CIFAR10 &  60000 &  1024 &  10 &      0.0 &     0.0 &     0.0 &   50.51 &          82.34 &          55.58 &            0.0 &  81.41 \\
\hline
\end{tabular}%
}
\caption{Sparsity results for all UCI and vision data sets, where $n$ denotes the data size, $p$ denotes the data dimension, and $k$ denotes the number of classes. Darker blue corresponds to higher nominal (DL) natural accuracy.}\label{tab:a4}
\end{table}

\end{appendices}








\end{document}